\documentclass[12pt]{l4dc2021} 

\title[Cautious Bayesian Optimization for Efficient and Scalable Policy Search]{Cautious Bayesian Optimization \\ for Efficient and Scalable Policy Search}
\usepackage{times}
\usepackage{algpseudocode,algorithm}
\usepackage{amsmath, amsfonts}
\usepackage{bm}
\usepackage{caption}
\usepackage{glossaries}
\usepackage{graphicx}
\usepackage{lipsum}
\usepackage{mathtools}
\usepackage{microtype}      
\usepackage{multicol}
\usepackage{todonotes}
\usepackage{url}
\usepackage{varwidth}
\usepackage{wrapfig}

\usepackage{pgfplots}
\usepackage{tikz}
\usetikzlibrary{intersections}
\usetikzlibrary{calc}
\usetikzlibrary{shapes.geometric}
\usetikzlibrary{decorations.markings}
\usetikzlibrary{decorations.pathmorphing}

\newacronym{acr:bo}{BO}{Bayesian optimization}
\newacronym{acr:crbo}{CRBO}{confidence region Bayesian optimization}
\newacronym{acr:dmp}{DMP}{dynamic movement primitive}
\newacronym{acr:ei}{EI}{expected improvement}
\newacronym{acr:ep}{EP}{expectation propagation}
\newacronym{acr:es}{ES}{entropy search}
\newacronym[firstplural=Gaussian processes (GPs)]{acr:gp}{GP}{Gaussian process}
\newacronym{acr:ir}{IR}{inference regret}
\newacronym{acr:nn}{NN}{neural network}
\newacronym{acr:pi}{PI}{probability of improvement}
\newacronym{acr:ps}{PS}{policy search}
\newacronym{acr:ppo}{PPO}{proximal policy optimization}
\newacronym{acr:rl}{RL}{reinforcement learning}
\newacronym{acr:se}{SE}{squared exponential}
\newacronym{acr:turbo}{TuRBO}{trust region Bayesian optimization}
\newacronym{acr:ucb}{UCB}{upper confidence bound}
\glsdisablehyper

\newcommand{\lfalgref}[1]{Alg.\,\ref{#1}}
\newcommand{\lfeqref}[1]{Eq.\,\eqref{#1}}
\newcommand{\lffigref}[1]{Fig.\,\ref{#1}}
\newcommand{\lfsecref}[1]{Sec.\,\ref{#1}}
\newcommand{\lftabref}[1]{Tab.\,\ref{#1}}

\definecolor{tableauC0}{rgb}{0.122, 0.467, 0.706}
\definecolor{tableauC1}{rgb}{1.000, 0.500, 0.055}
\definecolor{tableauC2}{rgb}{0.172, 0.627, 0.172}
\definecolor{tableauC3}{rgb}{0.839, 0.153, 0.157}
\definecolor{tableauC4}{rgb}{0.578, 0.402, 0.738}
\definecolor{tableauC5}{rgb}{0.547, 0.336, 0.293}
\definecolor{tableauC6}{rgb}{0.887, 0.465, 0.759}
\definecolor{tableauC7}{rgb}{0.500, 0.500, 0.500}
\definecolor{tableauC8}{rgb}{0.734, 0.738, 0.133}

\newcommand{\datan}{\mathcal{D}_n}
\newcommand{\confregion}{\mathcal{C}_n}
\newcommand{\confregionapprox}{\tilde{\mathcal{C}}_n}
\newcommand{\btheta}{\bm{\theta}}
\newcommand{\state}{\bm{s}}
\newcommand{\action}{\bm{a}}

\newcommand{\avgreward}{\bar{R}_n}
\newcommand\norm[1]{\left\lVert#1\right\rVert}

\tikzset{triangleupsidedown/.style={regular polygon, regular polygon sides=3, rotate=180, minimum size=3*(#1-\pgflinewidth), inner sep=0pt, outer sep=1pt}, triangleupsidedown/.default={5pt}}

\tikzset{triangle/.style={regular polygon, regular polygon sides=3, inner sep=0pt, outer sep=0pt}, triangle/.default={5pt}}

\usepackage{array}
\newcolumntype{L}[1]{>{\raggedright\let\newline\\\arraybackslash\hspace{0pt}}m{#1}}
\newcolumntype{C}[1]{>{\centering\let\newline\\\arraybackslash\hspace{0pt}}m{#1}}
\newcolumntype{R}[1]{>{\raggedleft\let\newline\\\arraybackslash\hspace{0pt}}m{#1}}

\DeclareMathOperator*{\argmin}{argmin}
\DeclareMathOperator*{\argmax}{argmax}
\DeclarePairedDelimiter{\ceil}{\lceil}{\rceil}

\definecolor{tableaublue}{rgb}{0.122,0.467,0.706}
\definecolor{tableauorange}{rgb}{1.0, 0.5, 0.055}
\definecolor{tableaugreen}{rgb}{0.172, 0.627, 0.172}
\definecolor{tableaured}{rgb}{0.839, 0.153, 0.157}

\newcommand{\figurefontsize}{\footnotesize}

\newcommand{\distancebeforeheading}{\vspace{-3mm}}
\newcommand{\distanceafterheading}{\vspace{-2mm}}

\newboolean{arxiv} 
\setboolean{arxiv}{true} 

\author{%
 \Name{Lukas P. Fr\"ohlich\nametag{\thanks{Work done while at Bosch Center for Artificial Intelligence, Renningen, Germany.}}} \Email{lukasfro@ethz.ch} \\
 \Name{Melanie N. Zeilinger} \Email{mzeilinger@ethz.ch} \\
 \addr Institute for Dynamic Systems and Control, ETH Zurich, Switzerland
 \AND
 \Name{Edgar D. Klenske}\textsuperscript{\color{blue}{$\ast$}} \Email{edgar.klenske@gauss-ml.com}\\
 \addr Gauss Machine Learning, Stuttgart, Germany%
}

\begin{document}

\maketitle

\begin{abstract}%
Sample efficiency is one of the key factors when applying policy search to real-world problems.
In recent years, Bayesian Optimization~(BO) has become prominent in the field of robotics due to its sample efficiency and little prior knowledge needed.
However, one drawback of BO is its poor performance on high-dimensional search spaces as it focuses on global search.
In the policy search setting, local optimization is typically sufficient as initial policies are often available, e.g., via meta-learning, kinesthetic demonstrations or sim-to-real approaches.
In this paper, we propose to constrain the policy search space to a sublevel-set of the Bayesian surrogate model's predictive uncertainty.
This simple yet effective way of constraining the policy update enables BO to scale to high-dimensional spaces ($>$100) as well as reduces the risk of damaging the system.
We demonstrate the effectiveness of our approach on a wide range of problems, including a motor skills task, adapting deep RL agents to new reward signals and a sim-to-real task for an inverted pendulum system.%
\end{abstract}


\distancebeforeheading
\section{Introduction}
\distanceafterheading

Policy search has established itself as a powerful method for \gls{acr:rl} with numerous successful applications in robotics \citep{Deisenroth2013PolicySearchSurvey, Kober2013RoboticsSurvey, Stulp2013RobotSkillLearningSurvey}.
By using parameterized policies, the \gls{acr:rl} problem can be simplified, e.g., via problem-specific policy structures and/or informed parameter initialization from simulations or kinesthetic demonstrations \citep{Ijspeert2003DynamicMovementPrimitives, Khansari2011LearningStableNonlinearDS}.
For real-world applications, two key criteria have to be fulfilled by a policy search algorithm:
1) \textit{Sample efficiency} in terms of required experiments, as each interaction with a real system requires some form of human supervision and potentially inflicts wear and tear on the robot.
2) \textit{Safety considerations} to not harm the robot's environment and potentially itself.

In recent years, \gls{acr:bo} has seen rising interest in the robotics community as noted by recent surveys on policy search \citep{Sigaud2019PolicySearchOverview, Chatzilygeroudis2019PolicySearchHandfulTrials}.
The most important advantage of \gls{acr:bo} is its efficiency in terms of function evaluations due to the active choice of new sample points based on a Bayesian surrogate model of the objective.
Further, additional model knowledge can be included, e.g., via prior mean functions~\citep{Wilson2014TrajectoryBO}, task-specific kernels~\citep{Antonova2017DeepKernels}, multiple information sources from low-fidelity simulations and real experiments~\citep{Marco2017VirtualVsReal} as well as safety aspects~\citep{Berkenkamp2016ControllerOptimization}.
However, one drawback of \gls{acr:bo} is its focus on global search and as such \textit{``[BO] does not scale well to large policy parameter spaces''}~\citep[Sec.~3]{Sigaud2019PolicySearchOverview}.
For many applications, however, a global search of the parameters is not required as a good initial solution often exists, e.g., via simulations, demonstrations, model-based considerations or similar previously solved tasks.

In this paper, we introduce \textit{\gls{acr:crbo}}, a scalable extension to the well-known \gls{acr:bo} algorithm.
Starting with an initial guess for a policy, we propose to locally constrain the parameter space based on the Bayesian surrogate model's predictive uncertainty.
In particular, we only allow policies to be evaluated for which the surrogate model is confident about their outcome.
The benefits of this approach are numerous:
1)~We avoid over-exploration of the parameter space and thus retain the proven sample-efficiency of \gls{acr:bo}, even in high-dimensional settings.
2)~By staying close to previously evaluated policies, i.e., being cautious, we reduce the risk of damaging the system or its environment.
Additionally, we provide a worst-case analysis of the expected outcome for the policy to be evaluated.
3)~The hyperparameter governing the cautiousness of \gls{acr:crbo} has an intuitive explanation and is easy to tune in practice.
4)~Apart from Lipschitz continuity for the worst-case analysis, we make no assumptions about either the underlying system or the structure of the objective function as is common in order to scale \gls{acr:bo} to higher-dimensional parameter spaces.
The general applicability of \gls{acr:crbo} is demonstrated on a range of problems from robotics and continuous control.
We believe that \gls{acr:crbo} satisfies the need for an efficient and scalable policy search method that can additionally leverage the benefits from numerous works in the low-dimensional \gls{acr:bo} setting for robotics.

\distancebeforeheading
\section{Related Work}
\distanceafterheading

Scaling \gls{acr:bo} to high-dimensional parameter spaces is an active field of research.
A common assumption to alleviate the curse of dimensionality is that the objective function only varies along a low-dimensional subspace of the full parameter space \citep{Wang2016Rembo, Nayebi2019HesBo}, which in general is unknown.
Therefore, \cite{Wang2016Rembo} sample a random projection matrix and \cite{Nayebi2019HesBo} use a hashing function to construct an embedding that maps to the lower-dimensional space.
Both methods are limited to linear subspaces and the true effective dimensionality of the subspace is typically also unknown and thus treated as an additional hyperparameter.
Another approach is to assume an additive structure of the objective function
\citep{Kandasamy2015HighDimBO, Gardner2017DiscoveringAdditiveStructuresBO} and the key issue lies in inferring the a-priori unknown decomposition of optimization variables.
\cite{Kandasamy2015HighDimBO} show that knowledge of the true structure improves convergence for \gls{acr:bo}, however, in practice the best of randomly sampled decompositions is chosen.
If no such structure exists in the original objective, these approaches can lead to suboptimal solutions.
In the context of robotics, the objective function is typically known and the influence of certain optimization variables can be exploited.
\citet{Yuan2019WholeBodyBayesOpt} use domain specific knowledge of a whole-body-control formulation in order to identify a suitable subspace during optimization.
\citet{Froehlich2019domainSelectionBO} make use of a probabilistic dynamics model to sample approximate solutions and reduce the search space via principal component analysis.
However, this approach is limited to linear dynamics due to the use of a linear quadratic regulator.

Constraining the parameter space during optimization is not uncommon, for example by assuming unknown constraints \citep{HernandezLobato2015PESUnknownConstraints} or by the notion of a critical region of the search space \citep{Berkenkamp2016ControllerOptimization, Marco2019ClassifiedRegressionBO}.
Unlike these methods, which assume the constraints to be part of the optimization problem itself, in \gls{acr:crbo} the domain is constrained by means of the surrogate model's uncertainty.
However, the ideas from \cite{HernandezLobato2015PESUnknownConstraints} and \cite{Marco2019ClassifiedRegressionBO} can be included in our framework.
A discussion about the difference between \gls{acr:crbo} and the SafeOpt framework \citep{Berkenkamp2016ControllerOptimization} is provided in \lfsecref{sec:crbo}.
Adapting the search space during the optimization has also been proposed borefore, but both approaches are limited to rectangular domains and do not consider cautiousness when increasing the search space \citep{Froehlich2019domainSelectionBO, Ha2019UnknownSearchSpaceBO}.

While standard \gls{acr:bo} is typically used for global optimization, there have been efforts to combine it with the advantages of local optimization, e.g., faster convergence in the vicinity of optima.
\cite{Wabersich2016MixedGlobalLocalKernelBO} propose a non-stationary kernel function that exploits the locally convex region around local and global minima.
\cite{McLeod2018SwitchingGlobalLocalBO} introduced the idea to use a local optimizer once the objective's surrogate model is certain about the location of the global optimum, however, this approach is limited to the noiseless setting.
Recently, \cite{Eriksson2019TuRBO} proposed to use multiple, rectangular trust-regions that only locally approximate the objective function.
These trust-regions are then updated based on heuristics instead of considering cautiousness and next evaluation points are suggested by posterior sampling of the surrogate model.
Similarly, \cite{Akrour2017LocalBO} use a sampling based approach where a Gaussian search distribution is updated in an (approximately) optimal manner adhering to information-theoretic constraints.
However, tuning the constraint parameters is non-trivial such that a good exploration-exploitation trade-off is difficult to obtain.
In the experimental section, we compare \gls{acr:crbo} to both \gls{acr:turbo} \citep{Eriksson2019TuRBO} and Local \gls{acr:bo} \citep{Akrour2017LocalBO}.

\distancebeforeheading
\section{Preliminaries}
\distanceafterheading
\label{sec:Preliminaries}
In this section, we will formally introduce the policy search problem and briefly review \gls{acr:bo} as a solution strategy for the resulting optimization problem.
The goal of reinforcement learning is to find a policy $\pi: \mathbb{R}^{|\mathcal{S}|} \rightarrow \mathbb{R}^{|\mathcal{A}|}$ mapping states $\state \in \mathcal{S}$ to actions $\action = \pi(\state) \in \mathcal{A}$ that maximizes a predefined reward signal (see, e.g., \cite{Sutton1998Book}).
In the (episodic) policy search setting, the policy $\pi(\state ; \btheta)$ is parameterized by a vector $\btheta \in \bm{\Theta} \subseteq \mathbb{R}^d$, e.g., the weights of a \gls{acr:nn}.
The optimal policy is then found via maximization of an episode's expected return w.r.t. $\btheta$,
\begin{align}\begin{split}\label{eq:policy_search_problem}
\btheta^*  = \argmax_{\btheta \in \bm{\Theta}} J(\btheta), \quad 
\text{ with }
J(\btheta) = \mathbb{E}\left[ R(\bm{\tau}) | \bm{\theta} \right] = \int R(\bm{\tau}) p_{\bm{\theta}}(\bm{\tau}) d\bm{\tau},
\end{split}
\end{align}
where $\bm{\tau} = \{\state_t, \action_t \}_{t=1:T}$ denotes a state-action trajectory and the return is given by the sum of immediate rewards $R(\bm{\tau}) = \sum_t r(\state_t, \action_t)$ with $r: \mathcal{S} \times \mathcal{A} \rightarrow \mathbb{R}$.
The stochasticity of the return is the result from varying initial state conditions as well as the stochastic nature of the environment itself.
In this paper, we consider deterministic policies and address \eqref{eq:policy_search_problem} by means of \acrfull{acr:bo} based on noisy observations of the episodic returns.

\gls{acr:bo} is a method for global optimization of stochastic black-box functions designed to be highly sample efficient \citep{Frazier2018tutorialBO}.
The efficiency of \gls{acr:bo} stems from two main ingredients: 1)~a~Bayesian surrogate model that approximates the objective function based on noisy observations and 2) choosing the next evaluation point based on this model by means of an acquisition function $\alpha(\btheta)$.
A common choice for the Bayesian surrogate model are \glspl{acr:gp}, which define a prior distribution over functions, such that any finite number of function values are normally distributed with mean $\mu_0(\btheta)$ and covariance specified by the kernel function $k(\btheta, \btheta')$ for any $\btheta, \btheta' \in \bm{\Theta}$ (w.l.o.g. we assume $\mu_0(\btheta) \equiv 0$) \citep{Rasmussen2006Book}.
In the regression setting, the goal is to predict an unknown latent function \mbox{$J(\btheta): \bm{\Theta} \rightarrow \mathbb{R}$} based on noisy observations $\datan = \{(\btheta_i, y_i = J(\btheta_i) + \epsilon)\}_{i=1:n}$ with $\epsilon \sim \mathcal{N}(0, \sigma_\epsilon^2)$.
Given a Gaussian likelihood model, conditioning the \gls{acr:gp}'s prior distribution on data~$\datan$ leads to the closed-form posterior predictive distribution,
\begin{align}\label{eq:gp_predictive}
\mu_n(\btheta) = \bm{k}(\btheta)^\top \bm{K}^{-1} \bm{y}, \quad 
\sigma_n^2(\btheta) = k(\btheta, \btheta) - \bm{k}(\btheta)^\top \bm{K}^{-1} \bm{k}(\btheta),
\end{align}
at any $\btheta \in \bm{\Theta}$ with $[\bm{k}(\btheta)]_i = k(\btheta, \btheta_i)$, $[\bm{K}]_{ij} = k(\btheta_i, \btheta_j) + \delta_{ij}\sigma_\epsilon^2$, $[\bm{y}]_i = y_i$ and $\delta_{ij}$ denotes the Kronecker delta.
Typical choices for the kernel function in the context of \gls{acr:bo} are the \gls{acr:se} and Mat\'ern kernels \citep[Chapter 4]{Rasmussen2006Book}.

Given the probabilistic model of the objective function, the next evaluation point is chosen by maximizing the acquisition function $\alpha(\btheta)$ quantifying the exploration-exploitation trade-off between regions with large predicted values (exploitation) and regions of high uncertainty (exploration).
Numerous acquisition functions have been proposed in the literature, e.g., \gls{acr:ucb} \citep{Cox1992UpperConfidenceBound}, \gls{acr:ei} \citep{Mockus1975ExpectedImprovement} as well as information-theoretic variants \citep{Hennig2012EntropySearch, HernandezLobato2014PredictiveEntropySearch, Wang2017MaxValueEntropySearch}.
One of the major challenges associated with \gls{acr:bo} is the optimization of the acquisition function, which can be non-convex.
However, unlike the original objective function, the acquisition function is cheap to evaluate and often has analytical gradients such that local solvers can be employed.

\distancebeforeheading
\section{Confidence Region Bayesian Optimization}\label{sec:crbo}
\distanceafterheading

In standard \acrfull{acr:bo}, the user defines fixed box-constraints with lower and upper bounds $\btheta_l, \btheta_u$ for the optimization (policy) parameters $\btheta$.
If these bounds are chosen too tight, one might miss out on promising regions of the parameter space, but loosening the bounds leads to an increase of the search volume, which scales exponentially with the dimensionality of the parameters.
This issue is especially critical due to \gls{acr:bo}'s tendency of \textit{over-exploring} the domain boundaries at the beginning of the optimization process \citep{Siivola2018OverExplorationBO}.
In the context of policy search, this leads to many rollouts with poor performance, potentially damaging the system without gaining much information about the optimal parameters.

In practice, we often do not need to start an optimization process from scratch, but a suitable initial guess exists in many scenarios.
As a matter of fact, many approaches in robotics can provide such an initial solution: optimal control using model-based considerations, learning from demonstrations, meta-learning by previously solved tasks, transferring knowledge from simulation to hardware, etc.
Thus, the considered goal is to \textit{locally improve on the initial policy} instead of globally searching the parameter space as done in standard \gls{acr:bo}.
Akin to trust-region methods \citep[Chapter 4.4]{Nocedal2006NumericalOptimizationBook}, we constrain our next optimization iterate based on the agreement between the objective function and its surrogate model.
In other words, we only want to search for new parameters in regions where we are confident in our model, i.e., the \gls{acr:gp} approximating the objective.
Being a Bayesian model, \glspl{acr:gp} naturally provide this notion of confidence in terms of the predictive uncertainty given by \lfeqref{eq:gp_predictive}.
More formally, we propose to constrain the parameter space during optimization of the acquisition function such that
\begin{align}\label{eq:confidence_region}
\btheta_{n+1} = \argmax_{\btheta \in \confregion} \alpha(\btheta), \quad 
\text{ and } \confregion = \left\{ \btheta \in \mathbb{R}^d | \sigma_n(\btheta) \leq \gamma \sigma_f  \right\} \cap \bm{\Theta},
\end{align}
where $\confregion$ is referred to as the \textit{confidence region}, $\sigma_f$ denotes the signal standard deviation of the \gls{acr:gp}'s kernel and $\gamma \in (0, 1]$ is a tunable parameter, governing the effective size of $\confregion$.
For small values of $\gamma$, the confidence region is confined to be close to the data, whereas in the limit $\gamma = 1$, we obtain the standard \gls{acr:bo} formulation.
Based on this constraint for the parameter space, we name our method \textit{\acrfull{acr:crbo}}.
Throughout the paper, we keep $\gamma$ fixed during optimization, however, one could increase the parameter over time as to retain a theoretical no-regret bound akin to \cite{Srinivas2010UpperConfidenceBound}.

\begin{figure}[t]
 \figurefontsize
 \newcommand{\heightexampleviz}{0.23\linewidth} 
 \floatconts
 {fig:conf_region_visualization}
 {\setlength{\belowcaptionskip}{-15pt}
  \setlength{\abovecaptionskip}{-5pt}
  \caption{Two-dimensional example for the growing confidence region $\confregion$ (see \lfeqref{eq:confidence_region}) during the optimization process. The evaluated policies are denoted by the crosses, with red being the initial policy. In the beginning, CRBO follows the gradient direction for efficient exploration and then continues to explore and exploit local optima. Note that the algorithm does not get stuck in the local optimum in the upper left corner but continues to explore towards the global optimum instead.}}
 {%
  \subfigure[$n=3$]{\label{fig:conf_region_visualization_a}%
 \includegraphics[width=\heightexampleviz, height=\heightexampleviz, trim={2mm, 2mm, 2mm, 2mm}, clip]{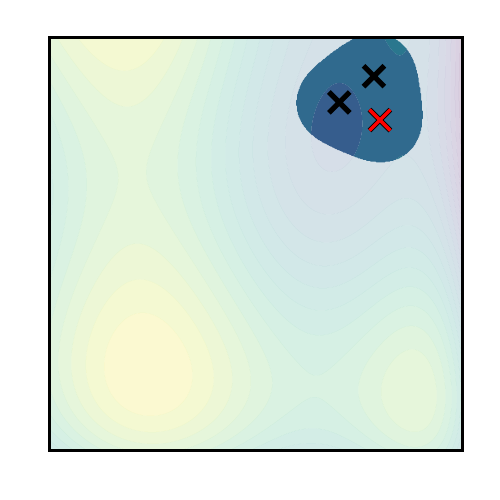}}\hfill
\subfigure[$n=10$]{\label{fig:conf_region_visualization_b}%
 \includegraphics[height=\heightexampleviz, width=\heightexampleviz, trim={2mm, 2mm, 2mm, 2mm}, clip]{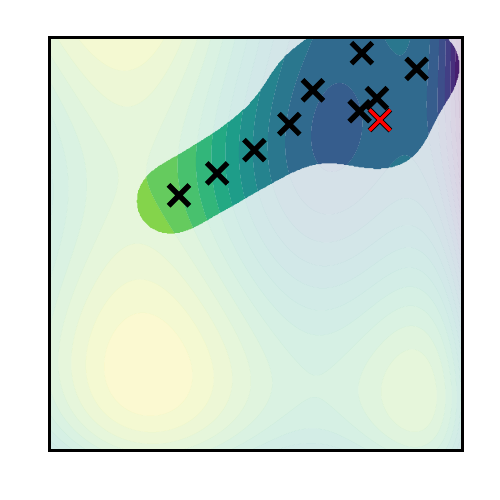}}\hfill
\subfigure[$n=20$]{\label{fig:conf_region_visualization_c}%
 \includegraphics[height=\heightexampleviz, width=\heightexampleviz, trim={2mm, 2mm, 2mm, 2mm}, clip]{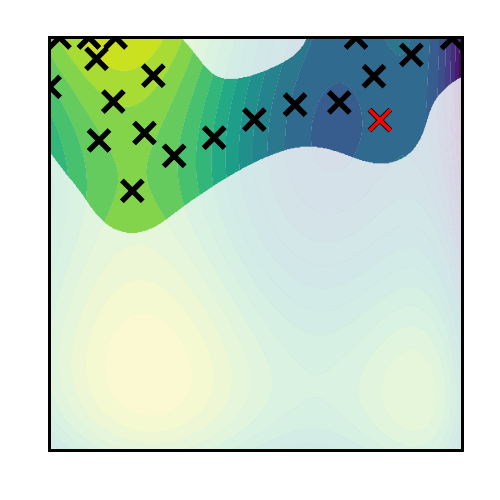}}\hfill
\subfigure[$n=50$]{\label{fig:conf_region_visualization_d}%
 \includegraphics[height=\heightexampleviz, width=\heightexampleviz, trim={2mm, 2mm, 2mm, 2mm}, clip]{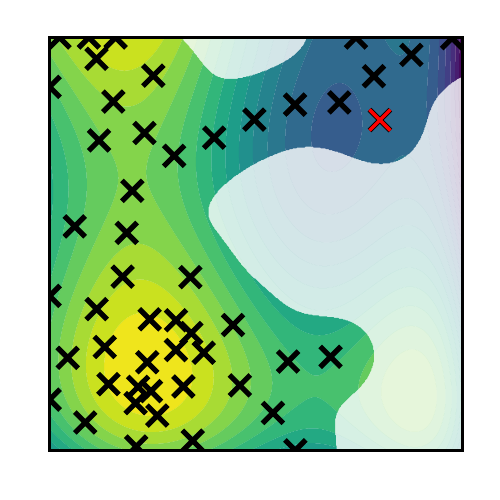}}
 }
\end{figure}

Note that the GP's predictive uncertainty depends on the observed data, such that the confidence region $\confregion$ can grow adaptively after each rollout of the system.
An exemplary run of \gls{acr:crbo} is depicted in \lffigref{fig:conf_region_visualization}.
After a local exploration phase around the initial policy (\lffigref{fig:conf_region_visualization_a}), \gls{acr:crbo} follows the gradient (\lffigref{fig:conf_region_visualization_b}) where the relative stepsize is governed by the confidence parameter $\gamma$.
Especially this greedy behavior in the beginning of the optimization is one of the key aspects that makes \gls{acr:crbo} highly efficient.
Once the algorithm finds a local optimum, it explores the surrounding region (\lffigref{fig:conf_region_visualization_c}).
Note that \gls{acr:crbo} does not get stuck in the local optimum, but instead continues exploring and finds the global optimum which it then exploits (\lffigref{fig:conf_region_visualization_d}).

One of the main benefits of \gls{acr:crbo} lies in its simplicity.
By only constraining the search space via \lfeqref{eq:confidence_region}, this approach is applicable to all acquisition functions, which enables \gls{acr:crbo} to be used in virtually any \gls{acr:bo} framework.
Further, the idea behind \gls{acr:crbo} can be used to improve other sophisticated variants such as multi-task \gls{acr:bo} \citep{Swersky2013MultiTaskBO},  multi-fidelity \gls{acr:bo} \citep{Marco2017VirtualVsReal} or manifold \gls{acr:bo} \citep{Jaquier2019RiemannBO}.

\distancebeforeheading
\subsection{Quantifying Cautiousness}
\distanceafterheading
Constraining the search space by \lfeqref{eq:confidence_region} implies that we stay close to previously evaluated parameters and, as a consequence, the parameter space is explored in a \textit{cautious} manner.
Consequently, we can estimate the worst-case outcome for the next rollout assuming the objective function is Lipschitz-continuous in the parameters, i.e., $\norm{J(\btheta_i) - J(\btheta_j)} \leq L_J \norm{\btheta_i - \btheta_j}$ for all $\btheta_i, \btheta_j \in \confregion$.
For this analysis, we first approximate the maximum distance of any evaluated policy to the boundary of the confidence region $\partial \confregion$ and second, we estimate the objective's Lipschitz constant $L_J$.

Note that if only given one data point $\btheta_0$, the confidence region is defined by a ball centered around $\btheta_0$ with radius $r_0  = \norm{\btheta - \btheta_0}$ determined by the condition $\sigma_0(\btheta; r_0) = \gamma \sigma_f$.
For the \gls{acr:se} kernel and $\sigma_{\epsilon}^2 \ll \sigma_f^2$, we can compute $r_0$ analytically; for other kernels, $r_0$ can be found numerically.
We then approximate the confidence region by a union of balls centered at the evaluated parameters,
\begin{align}\label{eq:confregion_approx}
\confregion \approx \confregionapprox = \cup_{i=1}^{n} \left\{ \btheta \in \bm{\Theta} | \norm{\btheta - \btheta_i} \leq r_0 \right\}, \quad
\text{ with } r_0^2 =  -2\ell^2 \log \sqrt{ 1 - \gamma^2}.
\end{align}
Thus, every possible next parameter $\btheta_{n+1}$ is within a distance of $r_0$ to at least one previously evaluated parameter in the dataset $\datan$.
Given this maximum step size and the bound on the change in output space via the Lipschitz constant, we can estimate the worst possible outcome for the next policy.
More specifically, the expected outcome of the policy parameterized by $\btheta_{n+1}$ is lower-bounded by
\begin{align}\label{eq:lipschitz_bound}
J(\btheta_{n+1}) \geq J(\btheta_{k}) - r_0(\gamma) \cdot L_J , \quad 
\text{ with } \btheta_k = \argmin_{\btheta_i \in \datan} \norm{\btheta_{n+1} - \btheta_i},
\end{align}
where we made the dependence of $r_0$ on the confidence region parameter $\gamma$ explicit.
In general, however, the true Lipschitz constant of the objective is not known a-priori and needs to be estimated from data, e.g., as the maximum of the norm of the \gls{acr:gp} mean gradient \mbox{$L_J \approx \max_{\btheta} \norm{\nabla \mu_n(\btheta)}$}.
While underestimating the Lipschitz constant can lead to a violation of \eqref{eq:lipschitz_bound}, this analysis offers valuable insight into the effect of the confidence parameter $\gamma$ on the cautiousness of \gls{acr:crbo}.
In practice, one could also adaptively tune the confidence parameter $\gamma$ such that a certain minimum performance threshold $J_{\text{min}}$ is not violated, i.e., $J(\btheta_{n+1}) \geq J_{\text{min}}$.

Note that the introduced concept of cautiousness is related albeit different from safety as presented by \cite{Berkenkamp2016ControllerOptimization}.
In particular, we do not assume an unknown safe set, e.g., defined by a performance threshold.
Further, by relaxing the strict safety assumption, the optimization of the acquisition function becomes scalable to high-dimensional spaces as we do not rely on (adaptive) grid search \citep{Berkenkamp2016ControllerOptimization, Duivenvoorden2017SafeOptParticleSwarmOpitmization}.
Lastly, \gls{acr:crbo} offers a parameter to directly tune the cautiousness whereas in the SafeOpt framework safety is only guaranteed under strict assumptions on the class of objective functions and no tunable parameter is provided.

\begin{figure}[t]
 \begin{minipage}{0.675\textwidth}
  \begin{algorithm}[H]
   \caption{Hit-and-Run sampler \citep{Belisle1993HitAndRun} used for optimization of the acquisition function in the confidence region $\confregion$.}
   \label{alg:hit_and_run}
   Choose starting point $\tilde{\btheta}_0 \in \mathcal{C}_n$ \\
   \For{k = 1:K} { 
    Generate uniformly random direction: $\bm{d} \in \mathbb{R}^d$ \\
    Find line set: $L_k = \mathcal{C}_n \cap \{ \btheta | \btheta = \btheta_{k-1} + \lambda \bm{d}, \lambda \in \mathbb{R} \}$ \\
    Sample next point $\tilde{\btheta}_{k} \sim \mathcal{U}(\btheta \in L_k)$
   }
   \textbf{return} Set of sample points $\{ \tilde{\btheta}_k \}_{k=1:K}$
  \end{algorithm}
 \end{minipage}\hfill
 \begin{minipage}{0.325\textwidth}
  \begin{figure}[H]
   \centering
%


\tikzset{%
    add/.style args={#1 and #2}{
            to path={%
                    ($(\tikztostart)!-#1!(\tikztotarget)$)--($(\tikztotarget)!-#2!(\tikztostart)$)%
                    \tikztonodes},add/.default={.2 and .2}}
}

\begin{tikzpicture}[scale=1.30]
    \draw [name path=curve, tableaublue, tension=0.8] plot [smooth cycle] coordinates {(0.5, 1.2)  (0.5, 2.3) (1.4, 2.6) (2.0, 2.0) (2.8, 1.3) (2.2, 0.3) (1.3, 1.0)};

    \coordinate (A) at (0.0, 0.0);
    \coordinate (B) at (2.0, 3.0);
    \coordinate (C) at (2.3, 1.2);
    \coordinate (D) at (0.5, 3.0);
    \coordinate (E) at (1.0, 1.5);

    \draw [name path=line1, add= 0.7 and 0.0, tableaured, densely dashed] (E) to (B);
    \draw [name intersections={of=line1 and curve, total=\n}]
    \foreach \i in {1,...,\n} {(intersection-\i) coordinate (curve-line1-intersection-\i)};
    \draw [tableaured, very thick] (curve-line1-intersection-1) -- node [](line1annotate) {} (curve-line1-intersection-2);

    \draw [name path=line2, add= 0.4 and 0.0, tableaugreen, densely dashed] (C) to (D);
    \draw [name intersections={of=line2 and curve, total=\n}]
    \foreach \i in {1,...,\n} {(intersection-\i) coordinate (curve-line2-intersection-\i)};
    \draw [tableaugreen, very thick] (curve-line2-intersection-1) -- node [](line2annotate) {}  (curve-line2-intersection-2);

    \node [tableaured](line1annotatetext) at (1.0, 0.5) {$L_1$};
    \draw [->,tableaured] (line1annotatetext.north) to [out=60,in=300] (line1annotate.south);

    \node [tableaugreen](line2annotatetext) at (2.5, 2.5) {$L_2$};
    \draw [->,tableaugreen] (line2annotatetext.south) to [out=250,in=30] (line2annotate.north);

    \path [name intersections={of=line1 and line2,by=line1-line2-intersection}];

    \node[inner sep=0pt] at (E) {$\bullet$};
    \node[left] at (E) {$\tilde{\btheta}_0$};

    \node[inner sep=0pt](F) at (line1-line2-intersection) {$\bullet$};
    \node[left] at (F) {$\tilde{\btheta}_1$};

    \node at (C) {$\bullet$};
    \node[below] at (C) {$\tilde{\btheta}_2$};

    \node[tableaublue] at (1.7, 1.1) {$\mathcal{C}_n$};

    %


\end{tikzpicture}

   \caption{Visualization of \lfalgref{alg:hit_and_run}.}
   \label{fig:hit_and_run}
  \end{figure}
 \end{minipage}
\vspace{-4mm}
\end{figure}

\distancebeforeheading
\subsection{Optimizing the Acquisition Function}
\distanceafterheading
Efficiently optimizing the acquisition function is crucial for the success of \gls{acr:bo}.
Even in rectangular optimization domains for standard \gls{acr:bo}, this is already a difficult problem due to the function's non-convex nature.
In practice, many \gls{acr:bo} toolboxes use numerical optimization methods, e.g., L-BFGS~\citep{Liu1989LBFGS}, with different starting points to find the acquisition function's optimum.
For \gls{acr:crbo} the problem is even more difficult, as the confidence region can become non-convex (\lffigref{fig:conf_region_visualization_d}).
We propose to evaluate the acquisition function at many locations which are uniformly distributed across $\confregion$ and then initialize a local optimizer at the location with the highest function value.
To deal with the nonlinearity of the constraints, we employ sequential quadratic programming.
While this procedure does not guarantee convergence to the global optimum, it works well in practice and is computationally efficient.

In order to initialize the local optimizer, we need to sample the confidence region uniformly, which for a non-convex bounded region is a hard problem in itself.
Simple methods, such as rejection sampling, do not scale well to the high-dimensional setting as the rejection rate increases exponentially with~$\dim{(\btheta)}$.
We therefore employ the so-called Hit-and-Run sampler \citep{Smith1984HitAndRun}, which works by iteratively choosing random one-dimensional subspaces on which rejection sampling is highly efficient.
Pseudo-code of the sampler is presented in \lfalgref{alg:hit_and_run} as well as an exemplary visualization of three consecutive samples in \lffigref{fig:hit_and_run}.
For more efficient coverage of the sample space, we use the fact that the observed data is already well distributed across the confidence region and thus start multiple parallel sampling chains from each of the data points.

\distancebeforeheading
\section{Experimental Results}\label{sec:results}
\distanceafterheading
We begin the experimental section with a comparison of standard \gls{acr:bo} and \gls{acr:crbo} on synthetic test functions (\lffigref{fig:withinmodel}) to emphasize the difference between a global and local exploration approach when a sufficiently good initial guess exists.
Further, we investigate a variety of control tasks.
In particular, we consider a complex motor-skill task where the initial policy stems from a (simulated) kinesthetic demonstration (\lffigref{fig:basketball}), different environments from the OpenAI gym for which pre-trained RL agents are adapted to new reward signals (\lffigref{fig:openai_results}),
and a sim-to-real task, where an initial policy is trained in simulation and fine-tuned on hardware (\lffigref{fig:furuta}).
Throughout all experiments, we use \gls{acr:ucb} as acquisition function for \gls{acr:crbo}, the Mat\'ern kernel as covariance function and we infer the MAP estimate of the \gls{acr:gp} hyperparameters after each iteration.
If not stated otherwise, each experiment was repeated 20 times.
The plots show the median as solid line and the interquartile range as shaded area.
More details about the parameters for each experiment are given in
\ifthenelse{\boolean{arxiv}}
{the Appendix \lfsecref{app:experimental_details}.}
{the accompanying tech report.}

\distancebeforeheading
\subsection{Low-dimensional Synthetic Functions}\label{sec:results_low_dim}
\distanceafterheading
\begin{figure}[t]
\figurefontsize
\newcommand{\heightsynthetic}{0.28\linewidth}
\newcommand{\widthsynthetic}{0.33\linewidth}
\newcommand{\heightexampleviz}{0.33\linewidth}
\newcommand{\bluelegendentry}{\raisebox{0pt}{\tikz{
   \draw[tableauC0!30!white,solid,fill=tableauC0!30!white,line width = 1.0pt](0.mm,0) rectangle (5.0mm,1.5mm);
   \node[draw,scale=0.4,circle,tableauC0,fill=tableauC0] at (2.5mm, 0.8mm){};
   \draw[-,tableauC0,solid,line width = 1.0pt](0.,0.8mm) -- (5.0mm,0.8mm)}}}
\newcommand{\redlegendentry}{\raisebox{0pt}{\tikz{
   \draw[tableauC3!30!white,solid,fill=tableauC3!30!white,line width = 1.0pt](0.mm,0) rectangle (5.0mm,1.5mm);
   \node[draw,scale=0.3, regular polygon, regular polygon sides=3, rotate=180, tableauC3,fill=tableauC3] at (2.5mm, 0.9mm){};
   \draw[-,tableauC3,solid,line width = 1.0pt](0.,0.8mm) -- (5.0mm,0.8mm)}}}
 \floatconts
 {fig:withinmodel}
 {\setlength{\belowcaptionskip}{-15pt}
  \setlength{\abovecaptionskip}{-5pt}
  \caption{Comparison of \gls{acr:crbo} (\protect\bluelegendentry) and standard \gls{acr:bo} (\protect\redlegendentry) on 2- and 5-dimensional synthetic benchmark functions, (b) and (c), respectively. For each experiment, the initial data point is chosen randomly with a distance of 0.3 to the global optimum.}}
 {%
  \subfigure[Visualization of a 2D function.]{\label{fig:within_model_c}%
   \begin{tikzpicture}
   \begin{axis}[
   x label style={at={(0.5,0.0)},anchor=center},
   xmin=-1.0,xmax=1.0,
   xlabel=Dimension 1,
   ylabel=Dimension 2,
   y label style={at={(0.10,0.5)},anchor=center},
   ymin=-1.0,ymax=1.0,
   major tick length=0.0cm,
   minor tick length = 0.0cm,
   tick pos=left,
   height=0.28\linewidth,
   width=0.30\linewidth,
   axis on top,
   ]
   \addplot[] graphics[xmin=-1.0,xmax=1.0,ymin=-1.0,ymax=1.0,] {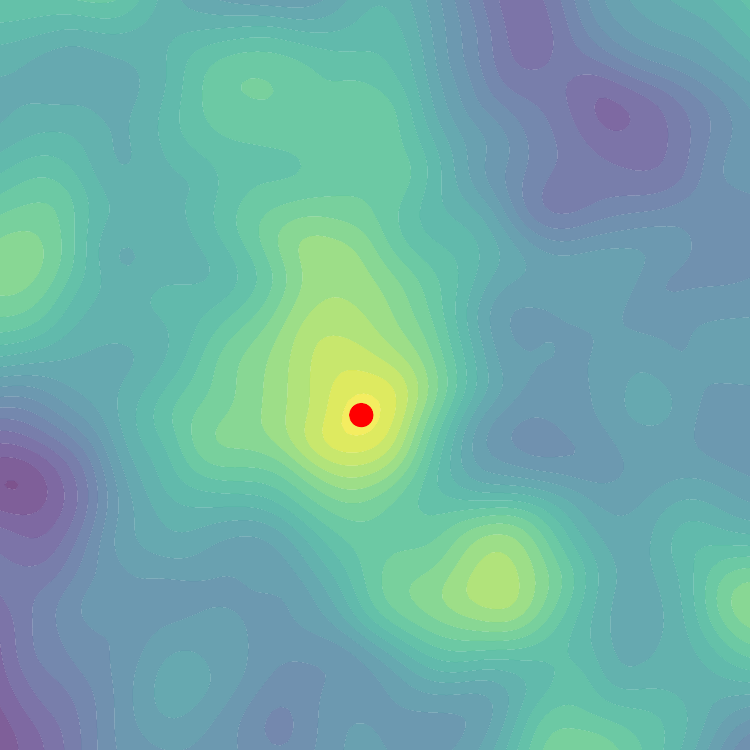};
  \end{axis}
  \end{tikzpicture}
}\hfill
  \subfigure[2D objective function.]{\label{fig:within_model_a}%
     \begin{tikzpicture}
  \begin{axis}[
  x label style={at={(0.5,0.0)},anchor=center},
  xmin=-2.5,xmax=51.5,
  xlabel=\# function evaluations,
  ylabel=Simple regret $s_n$,
     ymode = log,
  y label style={at={(0.08,0.5)},anchor=center},
  ymin=1.084e-6,ymax=4.529e0,
  major tick length=0.1cm,
  minor tick length = 0.05cm,
  tick pos=left,
  height=\heightsynthetic,
width=\widthsynthetic,
  axis on top,
     max space between ticks=20
  ]
  \addplot[] graphics[xmin=-2.5,xmax=51.5,ymin=1.084e-06,ymax=4.529e0,] {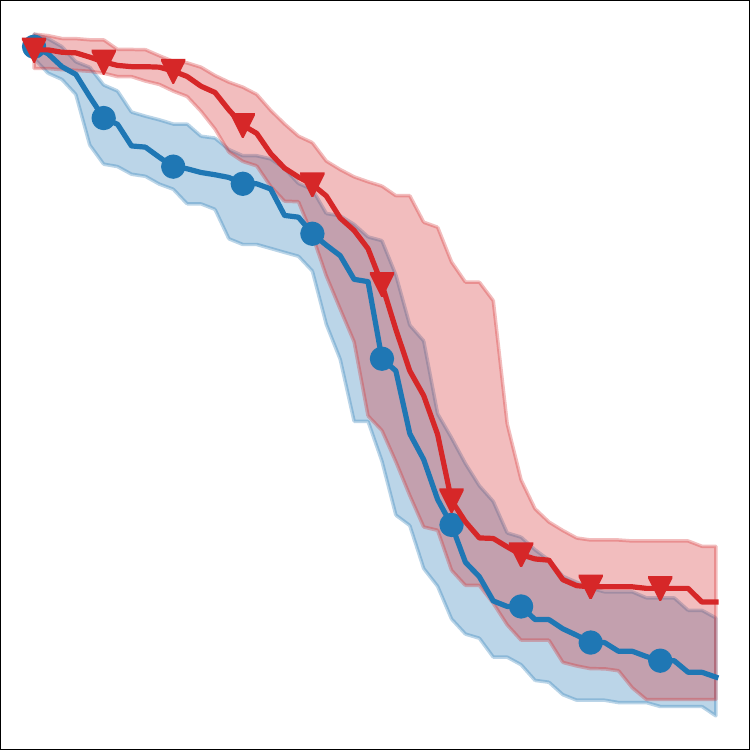};
  \end{axis}
  \end{tikzpicture} 
  }\hfill
  \subfigure[5D objective function.]{\label{fig:within_model_b}%
\begin{tikzpicture}
\begin{axis}[
x label style={at={(0.5,0.0)},anchor=center},
xmin=-10.0,xmax=208.9,
xlabel=\# function evaluations,
ylabel=Simple regret $s_n$,
ymode = log,
y label style={at={(0.08,0.5)},anchor=center},
ymin=4.290e-06,ymax=3.593e0,
major tick length=0.1cm,
minor tick length = 0.05cm,
tick pos=left,
height=\heightsynthetic,
width=\widthsynthetic,
axis on top,
max space between ticks=20
]
\addplot[] graphics[xmin=-10.0,xmax=208.9,ymin=4.290e-06,ymax=3.593e0,] {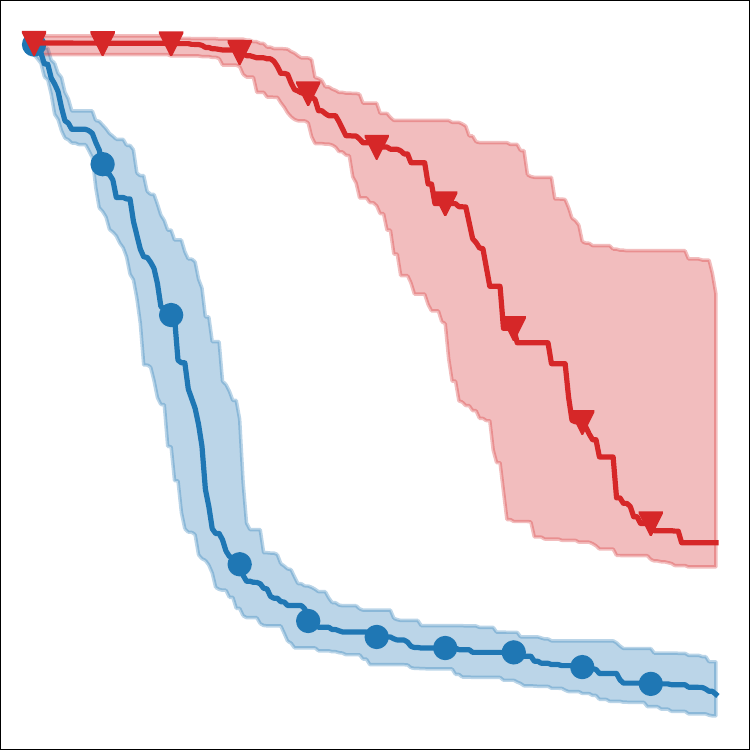};
\end{axis}
\end{tikzpicture}
}
 }
\end{figure}
The first experiment investigates the different exploration strategies of standard \gls{acr:bo} and \gls{acr:crbo}.
As objective functions, we draw 50 random samples from a \gls{acr:gp} with zero mean prior and a Mat\'ern kernel ($\sigma_f^2 = 1.0, \ell = 0.3$) on the domain $[-1, 1]^d$.
In order to avoid the optimum to be close to domain boundaries, we condition the \gls{acr:gp} on one training point in the center with $y = 3$.
A visualization of a resulting sample is shown in \lffigref{fig:within_model_c}.
For each objective, the initial data point was sampled randomly with a distance of 0.3 to the global optimum and the observation noise was fixed to $\sigma_\epsilon = 10^{-3}$.
The results in terms of simple regret, $s_n = \min_t |y_t - y^*|$, are depicted in \lffigref{fig:withinmodel}.
In the 2-dimensional setting, locally constraining the search space shows no significant advantage over the global exploration strategy.
As the search space is relatively small, standard \gls{acr:bo} quickly explores the domain and  converges to the global optimum.
Even for a moderately sized problem with 5 optimization parameters, however, the global exploration approach leads to slow convergence in the beginning and only after around 100 function evaluations, standard \gls{acr:bo} begins exploiting the region around the global optimum.
The local exploration approach is in stark contrast to this behavior.
Specifically, \gls{acr:crbo} follows the gradient direction from the beginning and thus requires significantly less function evaluations compared to standard \gls{acr:bo}.

\distancebeforeheading
\subsection{Policy Search Tasks}
\distanceafterheading
For the high-dimensional policy search tasks, we compare \gls{acr:crbo} to the following methods:
1)~\mbox{CMA-ES}~\citep{Hansen2001CMAES} as the gold standard for stochastic optimization, using the official Python implementation\footnote{\url{https://github.com/CMA-ES/pycma}},
2)~\gls{acr:turbo}~\citep{Eriksson2019TuRBO} as current state of the art for high-dimensional \gls{acr:bo}, using the provided Python implementation\footnote{\url{https://github.com/uber-research/TuRBO}}, and lastly
3)~Local~\gls{acr:bo}~\citep{Akrour2017LocalBO} as another \gls{acr:bo} approach based on locality, using a Matlab implementation.\footnote{Source code obtained from the authors via personal communication.}
For fair comparison, each algorithm starts with the same initial policy $\btheta_0$.

We measure an algorithm's performance with respect to two metrics: first, as measure for an algorithm's final performance, the expected return after optimization, $J(\btheta^*) = \mathbb{E}[R(\bm{\tau}) | \btheta^*]$, which we estimate using 10 rollouts.
Second, we consider the average of the observed returns during the optimization, $\avgreward = \frac{1}{n} \sum_{i=1:n} R(\bm{\tau}_i)$, akin to the cumulative regret in the \gls{acr:bo} literature, with $\bm{\tau}_i$ being the trajectory of the $i$-th episode.
While a good final performance $J(\btheta^*)$ is desired, $\avgreward$ essentially quantifies the cautiousness during the search for a good policy.
\input{figures/figure_basketball_new.tex}
To find the optimal value for each method's exploration parameter, we repeat the simulated experiments for different values of the exploration parameters in pre-defined ranges.
Clearly, this is not a viable approach for the hardware experiment.
For \gls{acr:crbo}, we therefore use the exploration parameter that performed best in the simulated experiments.
For a fair comparison, we tried three different values for the exploration parameter of CMA-ES in a truncated experiment, i.e., less episodes per experiment and fewer repetitions.
Only for the best performing parameter value we then conducted the full experiment.
To keep the figures interpretable, we only present a subset of the results in the main paper.
The full results are presented in
\ifthenelse{\boolean{arxiv}}
{the Appendix \lfsecref{app:additional_results}.}
{the accompanying tech report.}

\vspace{-2mm}
\paragraph{Learning from Demonstrations -- Basketball Task}\label{sec:results_basketball}
In this experiment, the goal is for a 7-DoF Kuka robotic arm to throw a basketball into a hoop as depicted in \lffigref{fig:basketball}.
Specifically, the reward is chosen as the minimum distance between the ball's trajectory and the center of the hoop.
With this experiment, we want to investigate the influence of each method's exploration parameter.
For CMA-ES and Local BO, this corresponds to the initial variance of the search distribution, for TuRBO this corresponds to the size of the initial rectangular trust region and for \gls{acr:crbo} we vary the confidence parameter $\gamma$ as to govern the confidence region's effective size.
For each method, we choose three different values ranging from small to large (exact values are given in the supplementary material).
The policy for the robot is parameterized by a \gls{acr:dmp}~\citep{Ijspeert2003DynamicMovementPrimitives} for each degree of freedom with $15$ basis functions each.
Additionally, the goal attractor for each joint angle was optimized, resulting in a total of $112$ free parameters.
The initial policy was learned by fitting the \glspl{acr:dmp} to a heuristically designed trajectory going through three via-points similar to a kinesthetic demonstration from a human.
For the simulation of the robot we use the PyRoboLearn framework \citep{Delhaisse2019pyrobolearn}.
The results are shown in \lffigref{fig:basketball}.
Common for all methods is that a large exploration factor leads to the smallest average observed return.
However, small exploration does not necessarily result in large average return as the algorithms can be stuck in local optima.
\gls{acr:crbo} is the only algorithm that shows monotonic improvement in terms of the average observed reward while all other methods are less cautious in the beginning of the optimization.
This is especially true for \gls{acr:turbo} due to its focus on global search.
Local \gls{acr:bo} demonstrates no significant improvement over the first 100 episodes which is in accordance with the originally presented results on a similarly complex task.
We therefore refrain from further comparison in other experiments.
In terms of final performance, \gls{acr:crbo} is slightly subpar compared to \gls{acr:turbo}, which can be seen as a trade-off for cautiousness.
\vspace{-2mm}
\paragraph{Adapting Deep RL Agents to New Rewards Signals}\label{sec:results_deeprl}
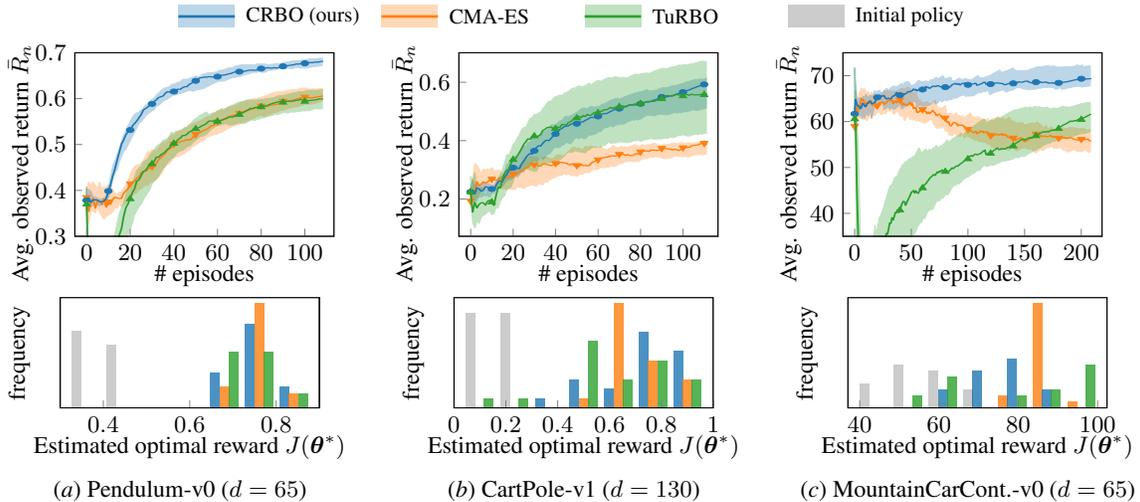
\begin{figure}[t]
\centering
\begin{tikzpicture}
\scriptsize
\def\barhalfheight{0.15}
\def\barwidth{0.75}
\def\dy{0.5}
\def\dxtext{1.0}
\def\dxbar{0.2}
\def\dx{2.7}

\newcommand{\LegendList}{
 0/tableauC0/solid/CRBO (ours), 
 1/tableauC1/solid/CMA-ES,
 2/tableauC2/solid/TuRBO
}

\foreach \i/\markercolor/\linestyle/\entry in \LegendList {
 \node[anchor=west] at (\dxtext + \i*\dx,0.0) {\entry};
 \fill [\markercolor!30!white] (\dxbar + \i*\dx,\barhalfheight) rectangle (\dxbar + \i*\dx + \barwidth, -\barhalfheight);
 \draw [-,\markercolor, \linestyle, line width = 1.0pt] (\dxbar + \i*\dx, 0.0) -- (\dxbar + \i*\dx + \barwidth, 0.0);
}

\newcommand{\LegendListTwo}{
 3/black/dashed/Initial policy
}

\foreach \i/\markercolor/\linestyle/\entry in \LegendListTwo {
 \node[anchor=west] at (\dxtext + \i*\dx,0.0) {\entry};
 \fill [\markercolor!20!white] (\dxbar + \i*\dx,\barhalfheight) rectangle (\dxbar + \i*\dx + \barwidth, -\barhalfheight);
}

\end{tikzpicture}

 \figurefontsize
 \newcommand{\heightopenai}{0.8\linewidth}
\begin{minipage}[t]{0.33\linewidth}
 \begin{tikzpicture}
 \begin{axis}[
 x label style={at={(0.5,0.0)},anchor=center},
 xmin=-5.4,xmax=113.4,
 xlabel=\# episodes,
 ylabel=Avg. observed return $\avgreward$,
 y label style={at={(0.15,0.4)},anchor=center},
 ymin=3.0e-01,ymax=7.0e-01,
 major tick length=0.1cm,
 minor tick length = 0.05cm,
 tick pos=left,
 height=\heightopenai,
 width=\linewidth,
 axis on top,
 max space between ticks=20
 ]
 \addplot[] graphics[xmin=-5.4,xmax=113.4,ymin=3.0e-01,ymax=7.0e-01,] {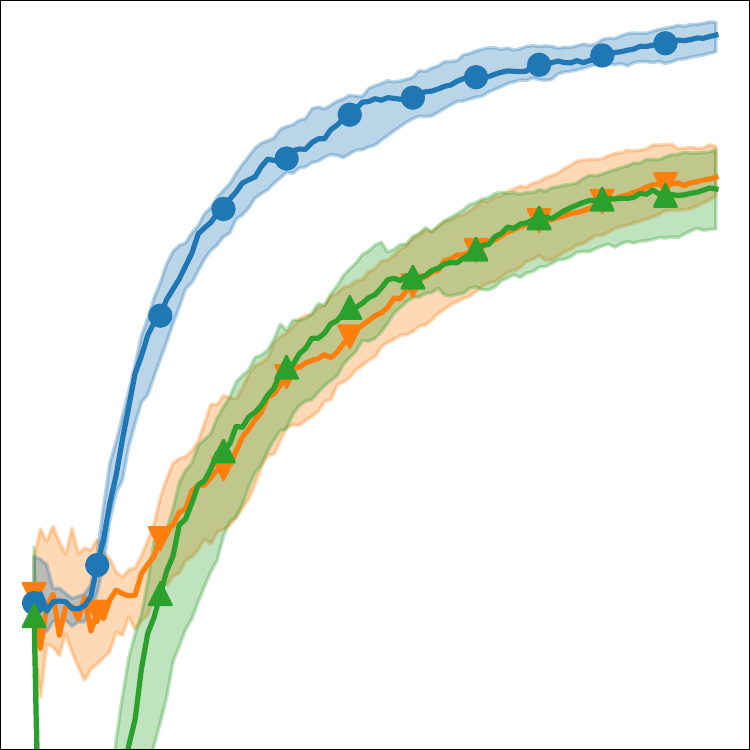};
 \end{axis}
 \end{tikzpicture}
\end{minipage}\hfill
\begin{minipage}[t]{0.33\linewidth}
 \begin{tikzpicture}
 \begin{axis}[
 x label style={at={(0.5,0.0)},anchor=center},
 xmin=-5.6,xmax=116.5,
 xlabel=\# episodes,
 ylabel=Avg. observed return $\avgreward$,
 y label style={at={(0.15,0.4)},anchor=center},
 ymin=7.252e-02,ymax=7.010e-01,
 major tick length=0.1cm,
 minor tick length = 0.05cm,
 tick pos=left,
 height=\heightopenai,
 width=\linewidth,
 axis on top,
 max space between ticks=20
 ]
 \addplot[] graphics[xmin=-5.6,xmax=116.5,ymin=7.252e-02,ymax=7.010e-01,] {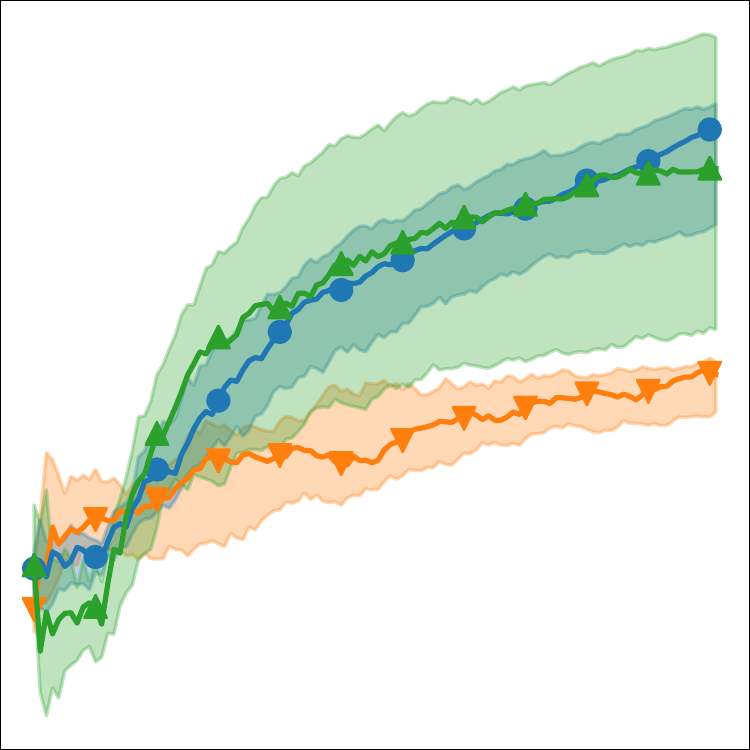};
 \end{axis}
 \end{tikzpicture}
\end{minipage}\hfill
\begin{minipage}[t]{0.33\linewidth}
\begin{tikzpicture}
\begin{axis}[
x label style={at={(0.5,0.0)},anchor=center},
xmin=-10.4,xmax=218.4,
ymin=3.5e+01,ymax=7.5e+01,
xlabel=\# episodes,
ylabel=Avg. observed return $\avgreward$,
y label style={at={(0.15,0.4)},anchor=center},
major tick length=0.1cm,
minor tick length = 0.05cm,
tick pos=left,
height=\heightopenai,
width=\linewidth,
axis on top,
max space between ticks=20
]
\addplot[] graphics[xmin=-10.4,xmax=218.4, ymin=3.5e+01,ymax=7.5e+01,] {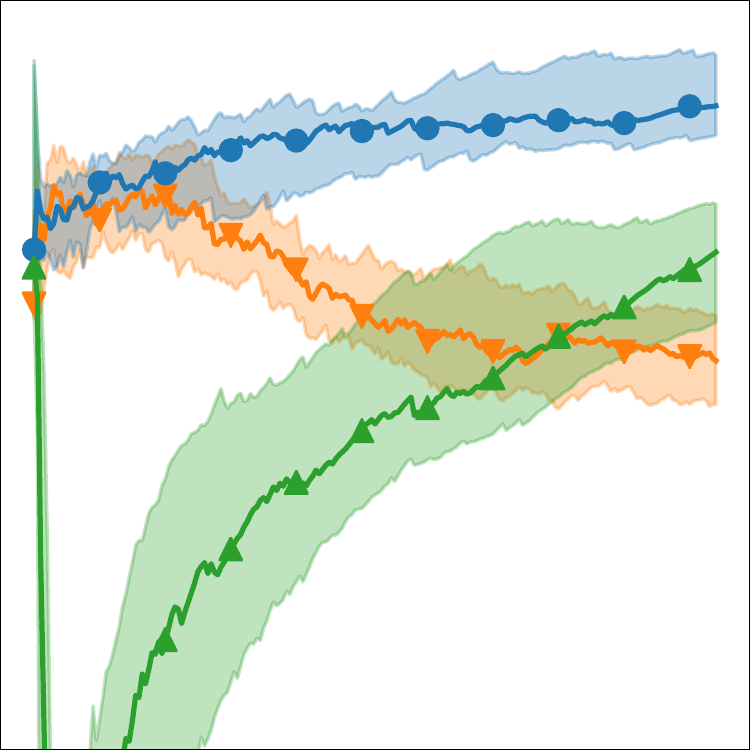};
\end{axis}
\end{tikzpicture}
\end{minipage}

 \newcommand{\heightfinalresult}{0.2\linewidth}
 \floatconts
 {fig:openai_results}
 {\setlength{\belowcaptionskip}{-15pt}
  \setlength{\abovecaptionskip}{-5pt}
  \caption{Results for OpenAI gym environments with adapted rewards signals.}}
 {%
  \subfigure[Pendulum-v0 ($d=65$)]{\label{fig:openai_pendulum}%
\begin{tikzpicture}
\begin{axis}[
x label style={at={(0.5,0.0)},anchor=center},
xmin=0.3,xmax=0.9,
xlabel=Estimated optimal reward $J(\btheta^*)$,
ylabel=frequency,
y label style={at={(0.2,0.5)},anchor=center},
ymin=0.0,ymax=1.078e+01,
major tick length=0.1cm,
minor tick length = 0.05cm,
tick pos=left,
ytick=\empty,
height=\heightfinalresult,
width=0.33\linewidth,
axis on top,
max space between ticks=20
]
\addplot[] graphics[xmin=0.3,xmax=0.9,ymin=0.0,ymax=1.078e+01,] {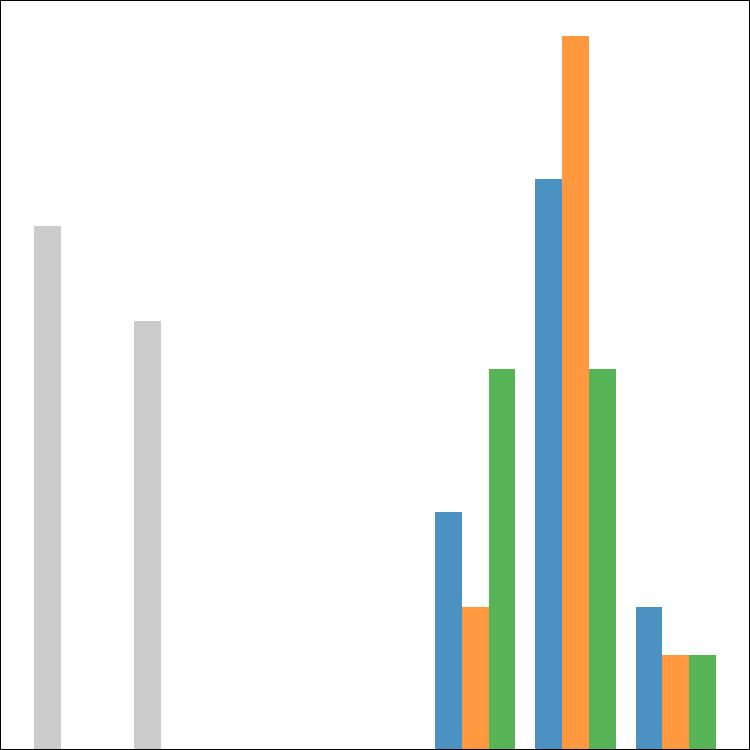};
\end{axis}
\end{tikzpicture}
  }\hfill
  \subfigure[CartPole-v1 ($d=130$)]{\label{fig:openai_cartpole}%
 \begin{tikzpicture}
\begin{axis}[
x label style={at={(0.5,0.0)},anchor=center},
xmin=0.0,xmax=1.0,
ymin=0.0,ymax=7.0,
xlabel=Estimated optimal reward $J(\btheta^*)$,
ylabel=frequency,
y label style={at={(0.2,0.5)},anchor=center},
major tick length=0.1cm,
minor tick length = 0.05cm,
tick pos=left,
ytick=\empty,
height=\heightfinalresult,
width=0.33\linewidth,
axis on top,
max space between ticks=20
]
\addplot[] graphics[xmin=0.0,xmax=1.0,ymin=0.0,ymax=7.0,] {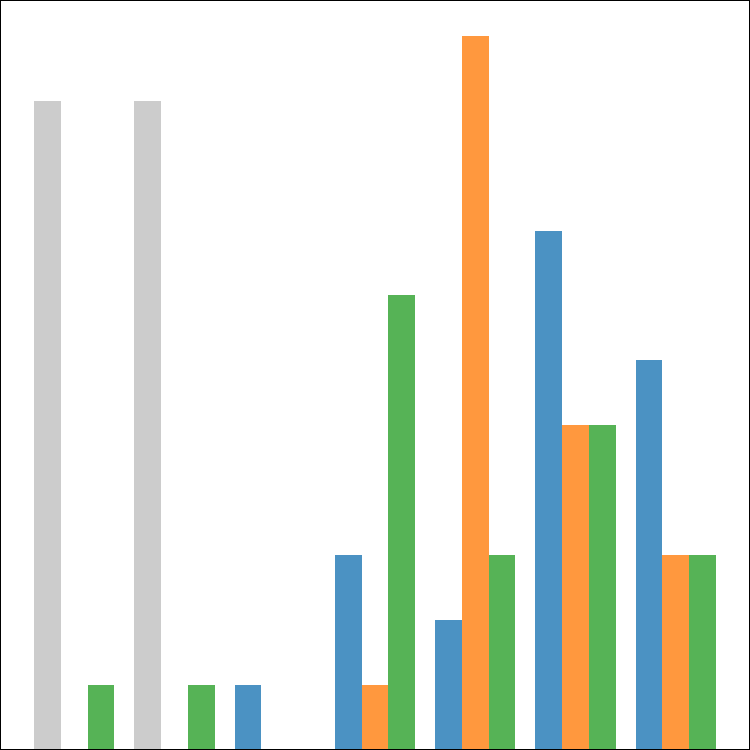};
\end{axis}
\end{tikzpicture}
  }\hfill
  \subfigure[MountainCarCont.-v0 ($d=65$)]{\label{fig:openai_mountaincar}%
\begin{tikzpicture}
\begin{axis}[
x label style={at={(0.5,0.0)},anchor=center},
xmin=37.0,xmax=102.4,
ymin=0.0,ymax=1.0,
xlabel=Estimated optimal reward $J(\btheta^*)$,
ylabel=frequency,
y label style={at={(0.2,0.5)},anchor=center},
major tick length=0.1cm,
minor tick length = 0.05cm,
tick pos=left,
ytick=\empty,
height=\heightfinalresult,
width=0.33\linewidth,
axis on top,
max space between ticks=20
]
\addplot[] graphics[xmin=37.0,xmax=102.4,ymin=0.0,ymax=1.0,] {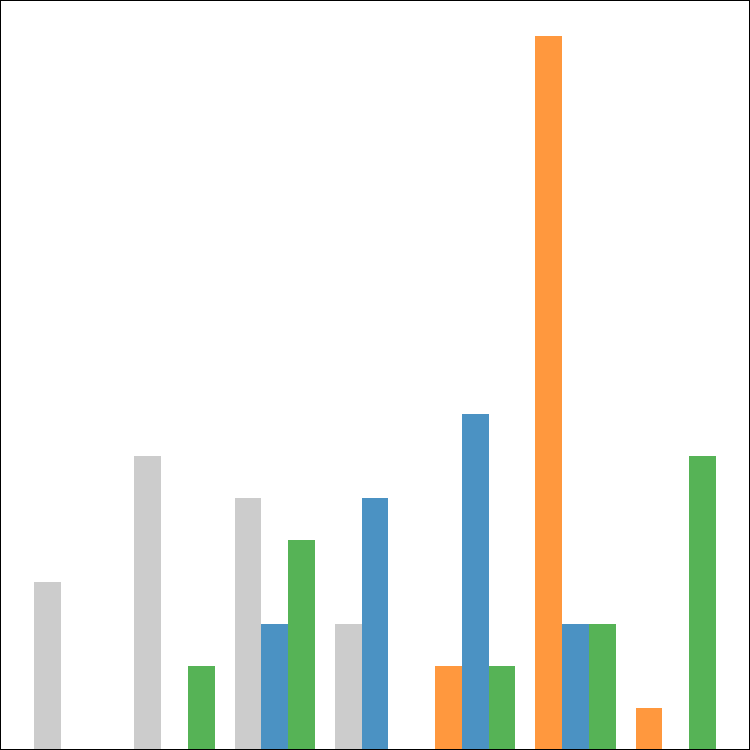};
\end{axis}
\end{tikzpicture}
  }%
 }
\end{figure}%
Recently, \cite{DeBruin2020FineTuningDeepRL} argued that deep RL algorithms are well suited to efficiently learn the feature extractor part of a \gls{acr:nn} policy, however, the so-called policy-head can be fine-tuned via gradient-free methods to further increase an agent's performance on the original reward signal.
In the next experiment, we investigate if an agent can quickly adapt to a different reward signal if the feature extractor part of the network is kept fixed.
Specifically, we use the agents trained with \gls{acr:ppo} \citep{Schulman2017PPO} provided by \citet{Raffin2018RLZoo}.
The policies are 2-layer \glspl{acr:nn} with $n_h = 64$ hidden units each, but we only optimize the weights of the linear mapping from the last hidden layer to the actions resulting in $(n_h + 1) \times | \mathcal{A} |$ optimization parameters.
We consider the following three environments from the OpenAI gym: Pendulum-v0, Cartpole-v1 and MountainCarContinuous-v0.
For each environment, we adjust the reward signal to fine-tune and improve the agent's behavior.
For the pendulum, we change the reward signal to $r_t = (1 - | \alpha_t | / 2^\circ) / T$ to promote a zero steady-state error of the pendulum's angle $\alpha$ at the end of an episode of length $T$.
The reward signal for the cartpole is changed to $r_t = (1 - |x_t|) / T$ with $x_t$ being the cart's position such that the system stays in the center.
Lastly, we penalize the action of the mountain car more strictly compared to the original reward to optimize for more energy efficient trajectories such that $r_t = -0.5 a_t^2$ and additional $+100$ reward if the car reaches the goal state.
The results are shown in \lffigref{fig:openai_results}.
Across all environments, \gls{acr:crbo} exhibits the best average observed return.
The final performance of all methods is similar for the Pendulum and CartPole environments, but for the MountainCar environment, the objective is extremely noisy which slows down \gls{acr:crbo}.
Note that CMA-ES inflates the covariance of its search distribution for increased exploration, leading to a decline in the average performance.
\gls{acr:turbo} evaluates many policies with poor performance in the beginning.
This is especially prominent for the Pendulum and MountainCar environments, for which the return is sensitive to changes in the parameter space.
\vspace{-2mm}
\paragraph{Sim-to-Real -- Stabilizing a Furuta Pendulum}\label{sec:results_furuta}
The next task considers the transfer of a policy trained in simulation to the hardware.
The Furuta pendulum consists of an unactuated pendulum attached to the end of a rotary arm that is actuated via a motor (see \lffigref{fig:furuta}); the dynamics of this system are similar to that of the cart-pole.
Similar to the previous experiment, we parameterize the policy by a 2-layer \gls{acr:nn} and optimize the weights of the last layer.
The initial policy is obtained by imitation learning in simulation with a stabilizing controller as expert policy using $30$ seconds of demonstration time and state/action pairs sampled at $250$ Hz.
The goal is to track a sinusoidal reference trajectory of the horizontal arm's angle while keeping the pendulum in the upright position.
The reward is given by $r_t = (1 - |\alpha_t - \alpha_{t, \text{ref}}| / 90^\circ) /T$ and an episode is terminated if the pendulum tips over or after $20$ seconds are reached.
We repeat the experiment 10 times for each method.
Due to the unpredictable behavior of \gls{acr:turbo} in the initial optimization phase, we do not include this method in the comparison.
On this task, \gls{acr:crbo} consistently outperforms CMA-ES both in terms of average and final performance by a large margin.
Trajectories of the arm's angle at various states of the learning progress are depicted in
\ifthenelse{\boolean{arxiv}}
{the Appendix \lfsecref{app:additional_results}.}
{the accompanying tech report.}


\begin{figure}[t]
\figurefontsize
 \newcommand{\heightaxis}{0.7\linewidth} 
 \newcommand{\bluelegendentry}{\raisebox{0pt}{\tikz{
   \draw[tableauC0!30!white,solid,fill=tableauC0!30!white,line width = 1.0pt](0.mm,0) rectangle (5.0mm,1.5mm);
   \node[draw,scale=0.3, regular polygon, regular polygon sides=3, rotate=180, tableauC0,fill=tableauC0] at (2.5mm, 0.9mm){};
   \draw[-,tableauC0,solid,line width = 1.0pt](0.,0.8mm) -- (5.0mm,0.8mm)}}}
\newcommand{\orangelegendentry}{\raisebox{0pt}{\tikz{
   \draw[tableauC1!30!white,solid,fill=tableauC1!30!white,line width = 1.0pt](0.mm,0) rectangle (5.0mm,1.5mm);
   \node[draw,scale=0.4,circle,tableauC1,fill=tableauC1] at (2.5mm, 0.8mm){};
   \draw[-,tableauC1,solid,line width = 1.0pt](0.,0.8mm) -- (5.0mm,0.8mm)}}}
\newcommand{\graylegendentry}{\raisebox{0pt}{\tikz{
   \draw[black!20!white,solid,fill=black!20!white,line width = 1.0pt](0.mm,0) rectangle (5.0mm,1.5mm);
}}}
\begin{minipage}[t]{0.4\linewidth}
\begin{tikzpicture}
\begin{axis}[
x label style={at={(0.5,0.0)},anchor=center},
xmin=-5.0,xmax=104.0,
ymin=5.0e-01,ymax=9.2e-01,
xlabel=\# episodes,
ylabel=Avg. observed return $\avgreward$,
y label style={at={(0.07 ,0.4)},anchor=center},
major tick length=0.1cm,
minor tick length = 0.05cm,
tick pos=left,
height=\heightaxis,
width=\linewidth,
axis on top,
max space between ticks=20
]
\addplot[] graphics[xmin=-5.0,xmax=104.0,ymin=5.0e-01,ymax=9.2e-01,] {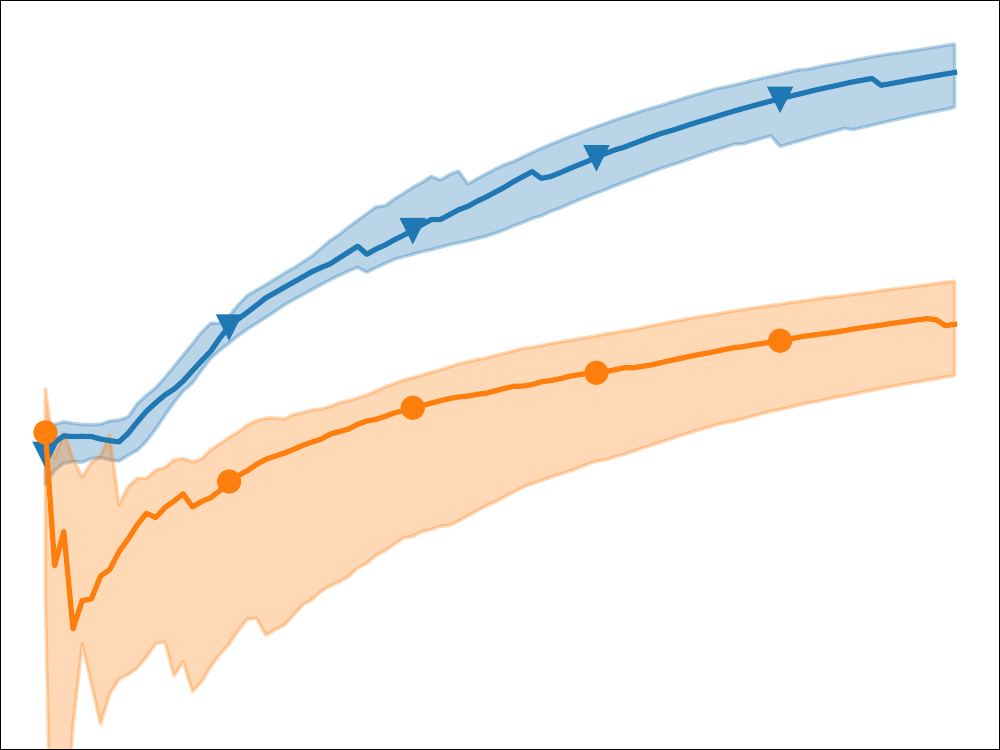};
\end{axis}
\end{tikzpicture}
\end{minipage}%
\begin{minipage}[t]{0.4\linewidth}
\begin{tikzpicture}
\begin{axis}[
x label style={at={(0.5,0.0)},anchor=center},
xmin=0.685,xmax=1.015,
ymin=0.0,ymax=1.0,
xlabel=Estimated optimal reward $J(\btheta^*)$,
ylabel=frequency,
y label style={at={(0.15,0.5)},anchor=center},
major tick length=0.1cm,
minor tick length = 0.05cm,
tick pos=left,
ytick=\empty,
height=\heightaxis,
width=\linewidth,
axis on top,
max space between ticks=20
]
\addplot[] graphics[xmin=0.685,xmax=1.015,ymin=0.0,ymax=1.0,] {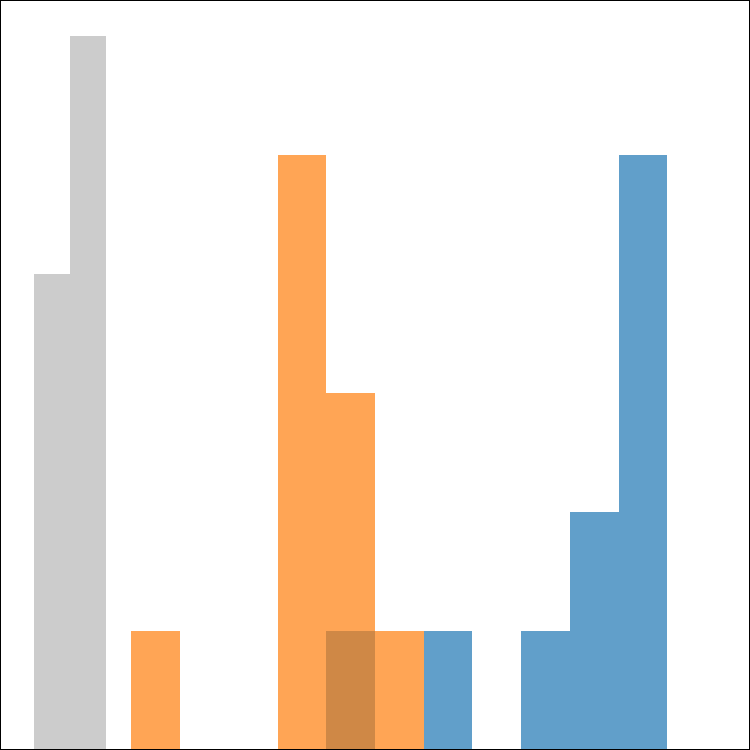};
\end{axis}
\end{tikzpicture}
\end{minipage}%
\begin{minipage}[t]{0.2\linewidth}
    \includegraphics[width=0.8\linewidth, trim={0mm, 0mm, 0mm, 0mm}, clip]{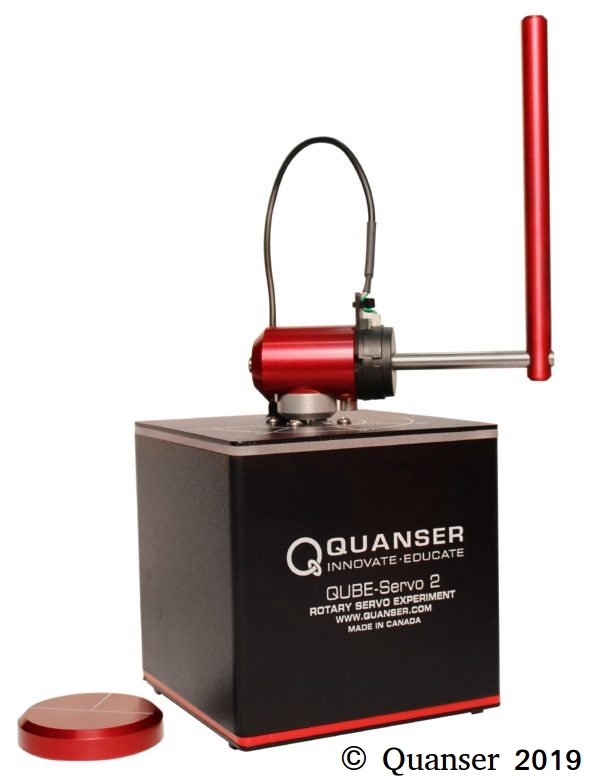}
\end{minipage}
\setlength{\belowcaptionskip}{-15pt}
\setlength{\abovecaptionskip}{-5pt}
\caption{Results for the the Furuta pendulum; \gls{acr:crbo} (\protect\bluelegendentry), CMA-ES (\protect\orangelegendentry), initial policy (\protect\graylegendentry).}
\label{fig:furuta}
\end{figure}

\vspace{-4mm}
\section{Conclusion}
\distanceafterheading
In this paper we have introduced \acrfull{acr:crbo}, a scalable extension to the well-known \gls{acr:bo} algorithm.
As such, \gls{acr:crbo} can be incorporated into a variety of existing \gls{acr:bo} methods.
The main goal of \gls{acr:crbo} is to retain the proven sample efficiency of standard \gls{acr:bo} by locally constraining the parameters which allows for optimization of higher-dimensional parameter spaces.
Specifically, we achieve this goal by constraining the search space to a sublevel-set of the surrogate model's predictive uncertainty.
The resulting algorithm can easily be tuned by an interpretable parameter that determines the cautiousness of the optimization.
We have demonstrated superior performance of \gls{acr:crbo} compared to state-of-the-art methods across many different tasks.


\clearpage
\acks{The authors thank Leonel Rozo and Philipp Hennig for their valuable feedback on this manuscript.
 The research of Melanie N. Zeilinger was supported by the Swiss National Science Foundation under grant no. PP00P2 157601/1.}

\bibliography{conf_names_long,library}  

\ifthenelse{\boolean{arxiv}}
{
 \clearpage
 \appendix
 \section{Additional Results}\label{app:additional_results}

\subsection{Computation Time}

We compare the time it takes for the different methods to provide a new evaluation point.
For \gls{acr:crbo} this corresponds to optimizing the acquisition function, for \gls{acr:turbo} it requires sampling from the \gls{acr:gp} posterior and finding the maximum, and for Local~BO it involves the optimization of a dual problem and locally sampling from the \gls{acr:gp} posterior.
We do not compare against CMA-ES here, as this method requires only sampling from a multivariate Gaussian search distribution whose mean and covariance matrix are updated via analytical equations every few rollouts, which of course requires significantly less computational effort compared with other methods.
We report the average (and standard deviation) compute time from an optimization run for the basketball task in \lftabref{tab:computation_time}.
The experiments were run on an Intel~Core~i5-7300~CPU~@~2.60GHz.

\begin{table}[h!]
 \centering
 \caption{Computation time (in sec) to provide a new evaluation point.}
 \label{tab:computation_time}
 \begin{tabular}{c|ccc}
  & CRBO &  TuRBO & LocalBO \\ \hline
  Duration  &  $0.501 \pm 0.391$      &  $2.618 \pm 0.170$       & $9.495 \pm 0.383$
 \end{tabular}
\end{table}

\subsection{Exploration Parameter Study}

As described in the main paper, we perform a study to choose the exploration parameter for the simulated experiments.
Here, we present all results of this study.
The values for the exploration parameters presented in the main paper are given in Section~\ref{app:experiment_details} of the supplementary material. 

\paragraph{Basketball} 
See \lffigref{fig:basketball_supplementary} for the results.
Note that for this particular task, CRBO is very robust with respect to the choice of the exploration parameter.
Further, CRBO is the only method that shows monotonic improvement across all values for the exploration parameter, whereas large exploration leads to drastic failure for all other methods.


\begin{figure}[]
 \newcommand{\heightopenai}{1.0\linewidth}
 \figurefontsize
 \begin{minipage}[t]{0.45\linewidth}
\begin{tikzpicture}
\begin{axis}[
x label style={at={(0.5,0.0)},anchor=center},
xmin=-5.5,xmax=115.5,
ymin=-5.339e-01,ymax=-1.943e-01,
xlabel=\# episodes,
ylabel=Avg. observed return $\avgreward$,
y label style={at={(0.02,0.5)},anchor=center},
major tick length=0.1cm,
minor tick length = 0.05cm,
tick pos=left,
height=\heightopenai,
width=1.0\linewidth,
axis on top,
max space between ticks=20
]
\addplot[] graphics[xmin=-5.5,xmax=115.5,ymin=-5.339e-01,ymax=-1.943e-01,] {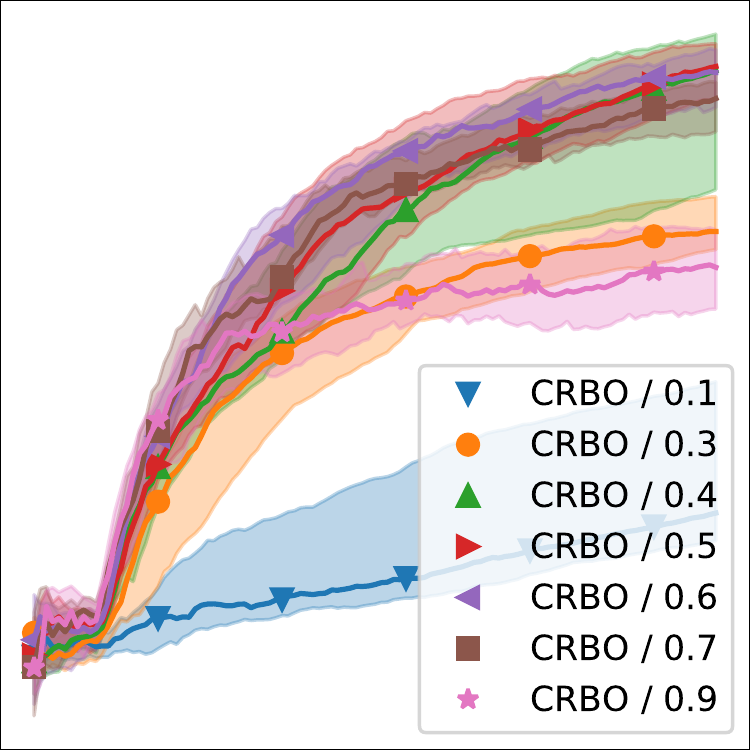};
\end{axis}
\end{tikzpicture}
 \end{minipage}\hfill
\begin{minipage}[t]{0.45\linewidth}
\begin{tikzpicture}
\begin{axis}[
x label style={at={(0.5,0.0)},anchor=center},
xmin=-5.5,xmax=115.5,
ymin=-1.142e-0,ymax=-2.765e-01,
xlabel=\# episodes,
ylabel=Avg. observed return $\avgreward$,
y label style={at={(0.02,0.5)},anchor=center},
major tick length=0.1cm,
minor tick length = 0.05cm,
tick pos=left,
height=\heightopenai,
width=1.0\linewidth,
axis on top,
max space between ticks=20
]
\addplot[] graphics[xmin=-5.5,xmax=115.5,ymin=-1.142e-0,ymax=-2.765e-01,] {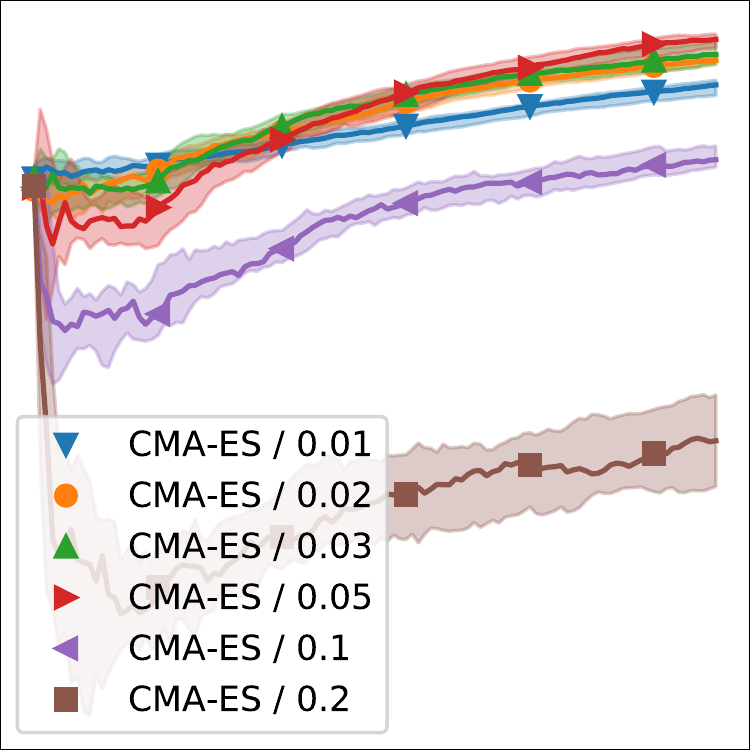};
\end{axis}
\end{tikzpicture}
\end{minipage}

 \begin{minipage}[t]{0.45\linewidth}
\begin{tikzpicture}
\begin{axis}[
x label style={at={(0.5,0.0)},anchor=center},
xmin=-5.5,xmax=115.5,
ymin=-1.574e-0,ymax=-2.468e-01,
xlabel=\# episodes,
ylabel=Avg. observed return $\avgreward$,
y label style={at={(0.02,0.5)},anchor=center},
major tick length=0.1cm,
minor tick length = 0.05cm,
tick pos=left,
height=\heightopenai,
width=1.0\linewidth,
axis on top,
max space between ticks=20
]
\addplot[] graphics[xmin=-5.5,xmax=115.5,ymin=-1.574e-0,ymax=-2.468e-01,] {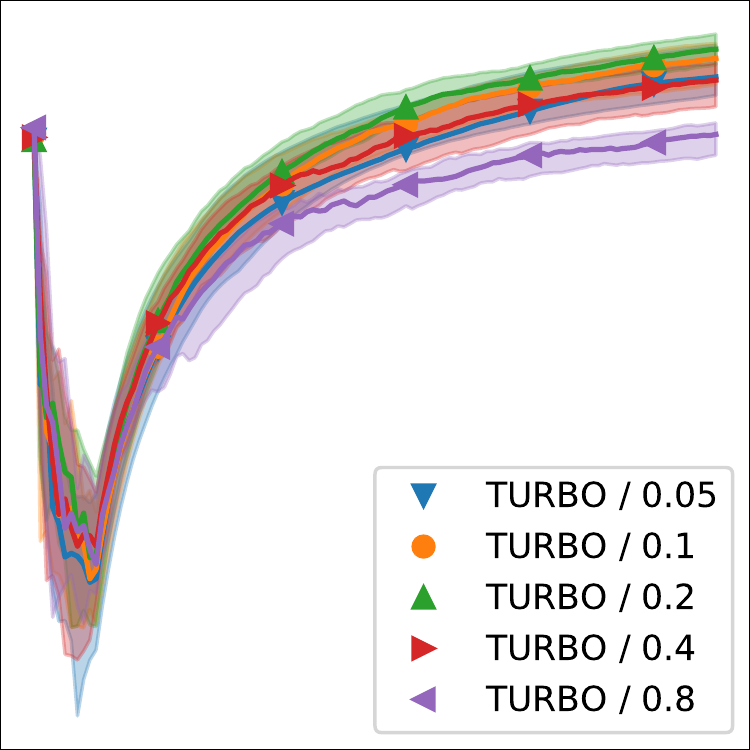};
\end{axis}
\end{tikzpicture}
\end{minipage}\hfill
\begin{minipage}[t]{0.45\linewidth}
\begin{tikzpicture}
\begin{axis}[
x label style={at={(0.5,0.0)},anchor=center},
xmin=-5.5,xmax=115.5,
ymin=-7.203e-01,ymax=-3.843e-01,
xlabel=\# episodes,
ylabel=Avg. observed return $\avgreward$,
y label style={at={(0.02,0.5)},anchor=center},
major tick length=0.1cm,
minor tick length = 0.05cm,
tick pos=left,
height=\heightopenai,
width=1.0\linewidth,
axis on top,
max space between ticks=20
]
\addplot[] graphics[xmin=-5.5,xmax=115.5,ymin=-7.203e-01,ymax=-3.843e-01,] {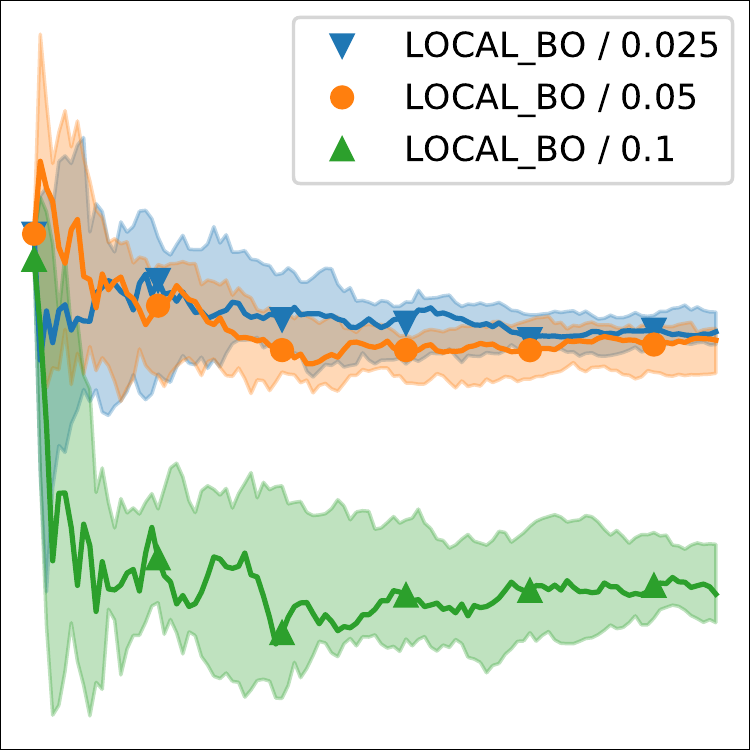};
\end{axis}
\end{tikzpicture}
\end{minipage}
\caption{Results for the Basketball Task using different values for each method's exploration parameters.}
\label{fig:basketball_supplementary}
\end{figure}

\paragraph{OpenAI} 
See \lffigref{fig:openai_supplementary} for the results.
In some cases, we chose to present the results for parameter values that are sub-optimal in terms of the average observed return, if in turn the final performance was better, e.g., for CMA-ES on the Cartpole environment (center column, center row).
Here, $\sigma_0 = 0.01$ is better in terms of the average observed return, however, the final performance is significantly better for $\sigma_0 = 0.02$, which we present in the main paper.
Further note that for the MountainCar environment (right column), we only use about 110 episodes for this study due to computational reasons, whereas in the main paper we use more than 200 episodes.


\begin{figure}[]
\newcommand{\heightopenai}{1.0\linewidth}
\figurefontsize
\begin{minipage}[t]{0.33\linewidth}
\begin{tikzpicture}
\begin{axis}[
x label style={at={(0.5,0.0)},anchor=center},
xmin=-5.4,xmax=113.4,
xlabel=\# episodes,
ylabel=Avg. observed return $\avgreward$,
y label style={at={(0.10,0.5)},anchor=center},
ymin=3.0e-01,ymax=7.0e-01,
major tick length=0.1cm,
minor tick length = 0.05cm,
tick pos=left,
height=\heightopenai,
width=1.0\linewidth,
axis on top,
max space between ticks=20
]
\addplot[] graphics[xmin=-5.4,xmax=113.4,ymin=3.0e-01,ymax=7.0e-01,] {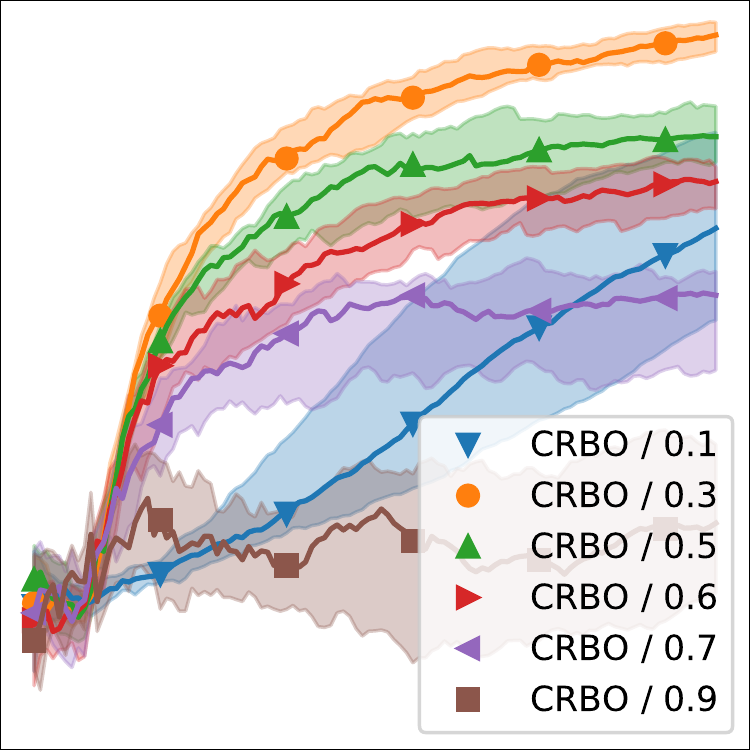};
\end{axis}
\end{tikzpicture}
\end{minipage}\hfill
\begin{minipage}[t]{0.33\linewidth}
 \begin{tikzpicture}
\begin{axis}[
x label style={at={(0.5,0.0)},anchor=center},
xmin=-5.6,xmax=116.5,
xlabel=\# episodes,
ylabel=Avg. observed return $\avgreward$,
y label style={at={(0.10,0.5)},anchor=center},
ymin=1.343e-01,ymax=6.529e-01,
major tick length=0.1cm,
minor tick length = 0.05cm,
tick pos=left,
height=\heightopenai,
width=1.0\linewidth,
axis on top,
max space between ticks=20
]
\addplot[] graphics[xmin=-5.6,xmax=116.5,ymin=1.343e-01,ymax=6.529e-01,] {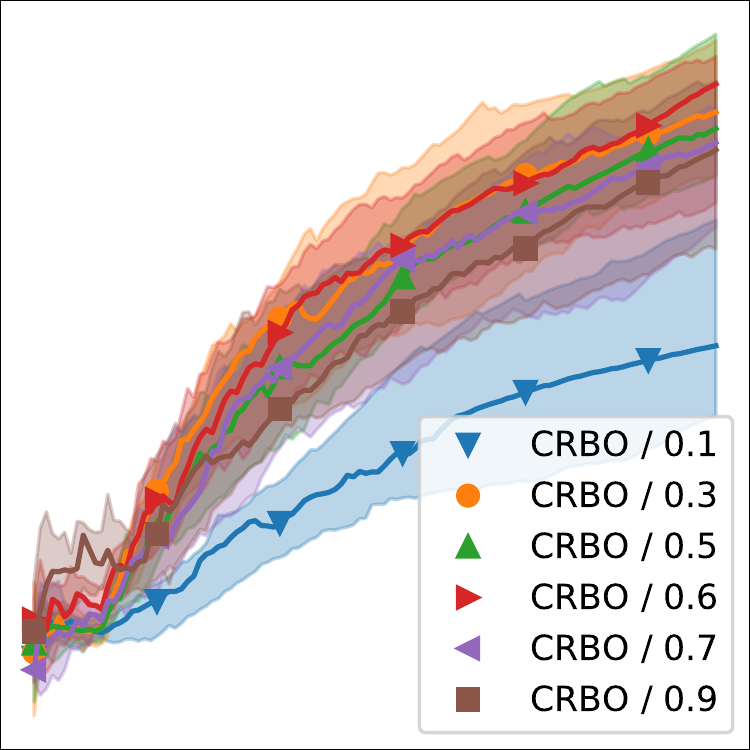};
\end{axis}
\end{tikzpicture}
\end{minipage}\hfill
\begin{minipage}[t]{0.33\linewidth}
\begin{tikzpicture}
\begin{axis}[
x label style={at={(0.5,0.0)},anchor=center},
xmin=-5.4,xmax=113.4,
ymin=3.5e+01,ymax=7.5e+01,
xlabel=\# episodes,
ylabel=Avg. observed return $\avgreward$,
y label style={at={(0.10,0.5)},anchor=center},
major tick length=0.1cm,
minor tick length = 0.05cm,
tick pos=left,
height=\heightopenai,
width=1.0\linewidth,
axis on top,
max space between ticks=20
]
\addplot[] graphics[xmin=-5.4,xmax=113.4, ymin=3.5e+01,ymax=7.5e+01,] {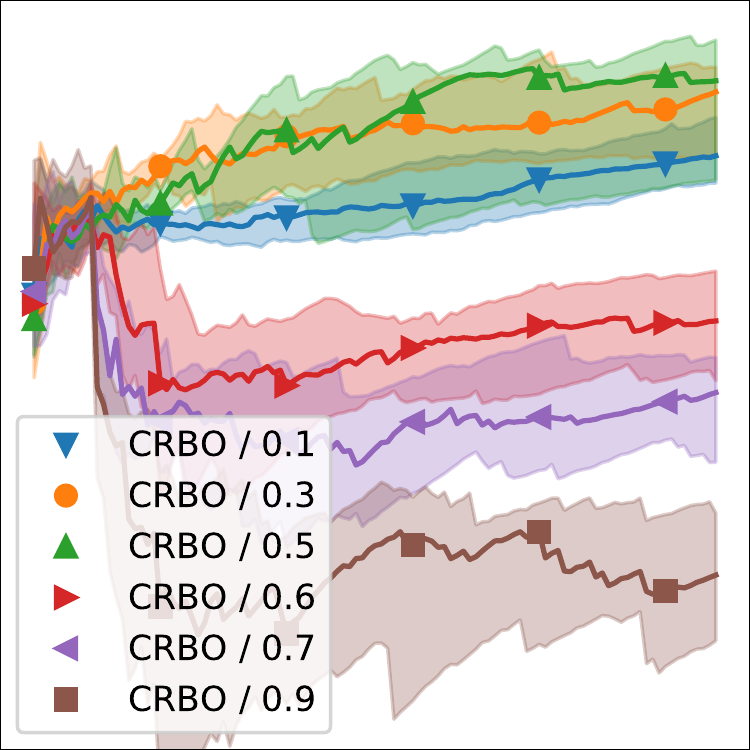};
\end{axis}
\end{tikzpicture}
\end{minipage}


\begin{minipage}[t]{0.33\linewidth}
 \begin{tikzpicture}
 \begin{axis}[
 x label style={at={(0.5,0.0)},anchor=center},
 xmin=-5.4,xmax=113.4,
 xlabel=\# episodes,
 ylabel=Avg. observed return $\avgreward$,
 y label style={at={(0.10,0.5)},anchor=center},
 ymin=1.269e-03,ymax=6.52e-01,
 major tick length=0.1cm,
 minor tick length = 0.05cm,
 tick pos=left,
 height=\heightopenai,
 width=1.0\linewidth,
 axis on top,
 max space between ticks=20
 ]
 \addplot[] graphics[xmin=-5.4,xmax=113.4, ymin=1.269e-03,ymax=6.52e-01,] {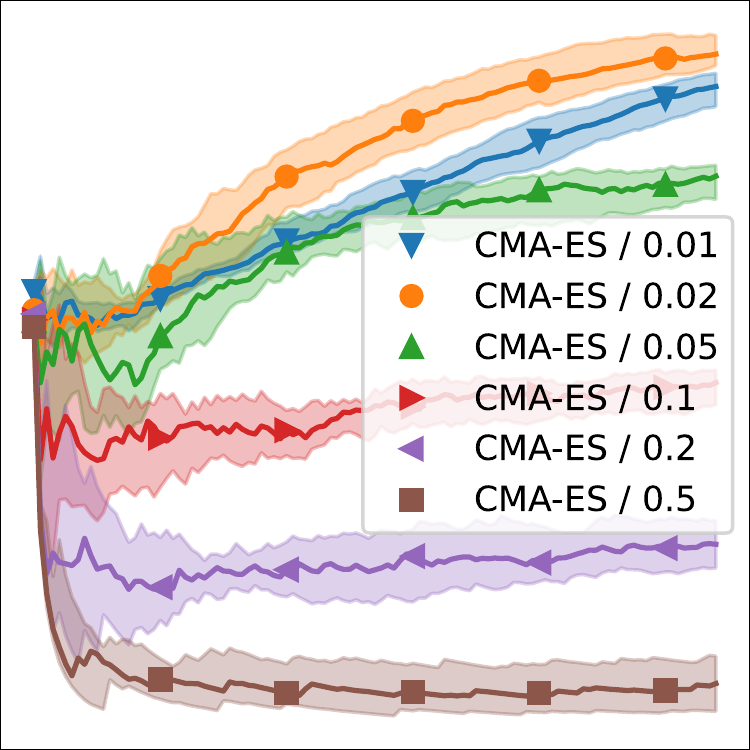};
 \end{axis}
 \end{tikzpicture}
\end{minipage}\hfill
\begin{minipage}[t]{0.33\linewidth}
 \begin{tikzpicture}
 \begin{axis}[
 x label style={at={(0.5,0.0)},anchor=center},
 xmin=-5.6,xmax=116.5,
 xlabel=\# episodes,
 ylabel=Avg. observed return $\avgreward$,
 y label style={at={(0.10,0.5)},anchor=center},
 ymin=6.484e-02,ymax=4.700e-01,
 major tick length=0.1cm,
 minor tick length = 0.05cm,
 tick pos=left,
 height=\heightopenai,
 width=1.0\linewidth,
 axis on top,
 max space between ticks=20
 ]
 \addplot[] graphics[xmin=-5.6,xmax=116.5, ymin=6.484e-02,ymax=4.700e-01,] {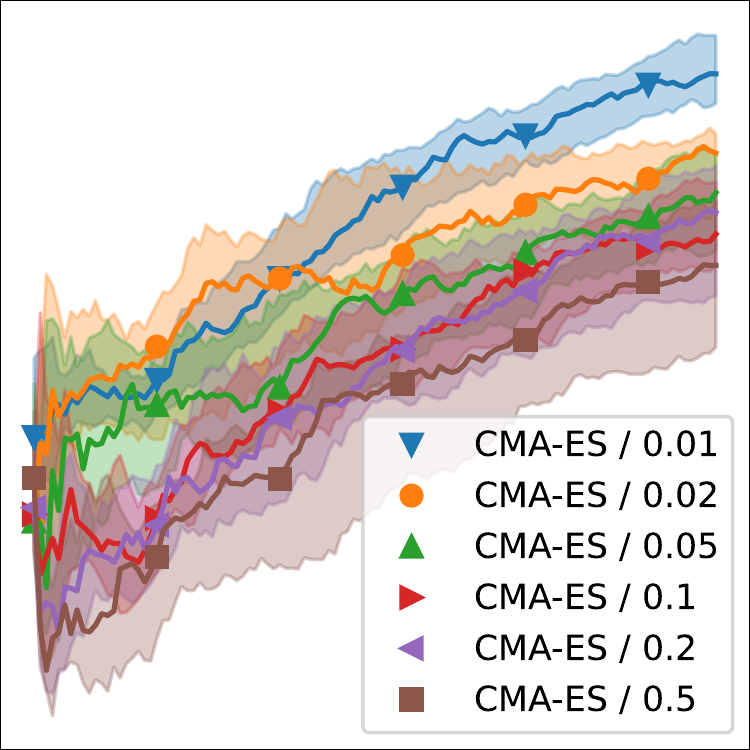};
 \end{axis}
 \end{tikzpicture}
\end{minipage}\hfill
\begin{minipage}[t]{0.33\linewidth}
 \begin{tikzpicture}
 \begin{axis}[
 x label style={at={(0.5,0.0)},anchor=center},
 xmin=-5.4,xmax=113.4,
 ymin=-2.5e+01,ymax=7.5e+01,
 xlabel=\# episodes,
 ylabel=Avg. observed return $\avgreward$,
 y label style={at={(0.10,0.5)},anchor=center},
 major tick length=0.1cm,
 minor tick length = 0.05cm,
 tick pos=left,
 height=\heightopenai,
 width=1.0\linewidth,
 axis on top,
 max space between ticks=20
 ]
 \addplot[] graphics[ xmin=-5.4,xmax=113.4,  ymin=-2.5e+01,ymax=7.5e+01,] {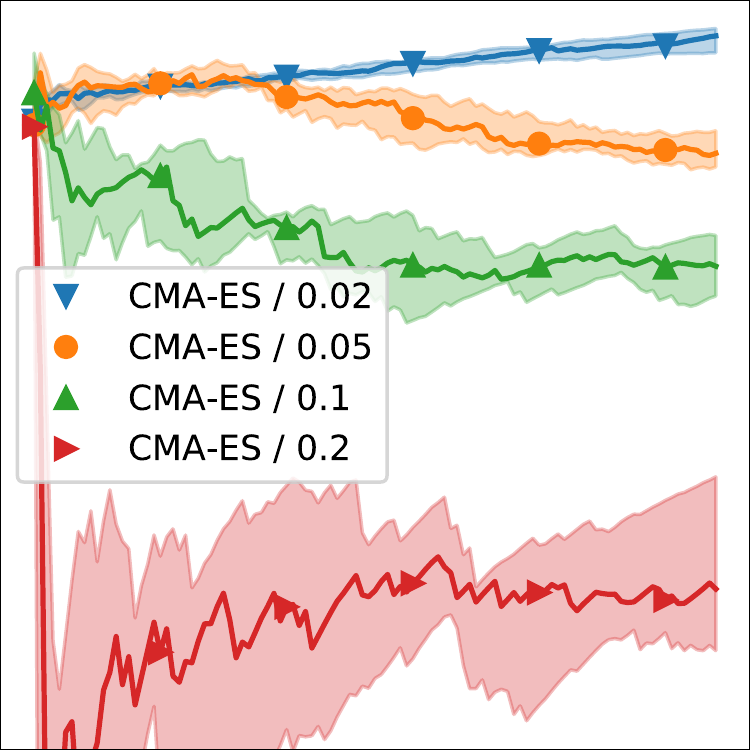};
 \end{axis}
 \end{tikzpicture}
\end{minipage}


\begin{minipage}[t]{0.33\linewidth}
 \begin{tikzpicture}
 \begin{axis}[
 x label style={at={(0.5,0.0)},anchor=center},
 xmin=-5.4,xmax=113.4,
 xlabel=\# episodes,
 ylabel=Avg. observed return $\avgreward$,
 y label style={at={(0.10,0.5)},anchor=center},
 ymin=2.201e-02,ymax=6.487e-01,
 major tick length=0.1cm,
 minor tick length = 0.05cm,
 tick pos=left,
 height=\heightopenai,
 width=1.0\linewidth,
 axis on top,
 max space between ticks=20
 ]
 \addplot[] graphics[xmin=-5.4,xmax=113.4,ymin=2.201e-02,ymax=6.487e-01,] {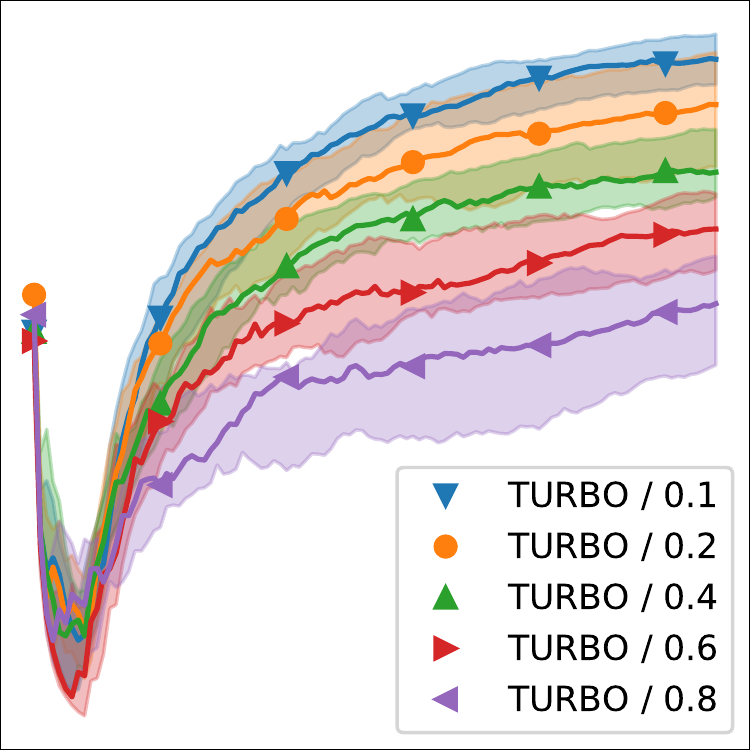};
 \end{axis}
 \end{tikzpicture}
\end{minipage}\hfill
\begin{minipage}[t]{0.33\linewidth}
 \begin{tikzpicture}
 \begin{axis}[
 x label style={at={(0.5,0.0)},anchor=center},
 xmin=-5.6,xmax=116.5,
 xlabel=\# episodes,
 ylabel=Avg. observed return $\avgreward$,
 y label style={at={(0.10,0.5)},anchor=center},
 ymin=6.157e-02,ymax=7.016e-01,
 major tick length=0.1cm,
 minor tick length = 0.05cm,
 tick pos=left,
 height=\heightopenai,
 width=1.0\linewidth,
 axis on top,
 max space between ticks=20
 ]
 \addplot[] graphics[xmin=-5.6,xmax=116.5, ymin=6.157e-02,ymax=7.016e-01,] {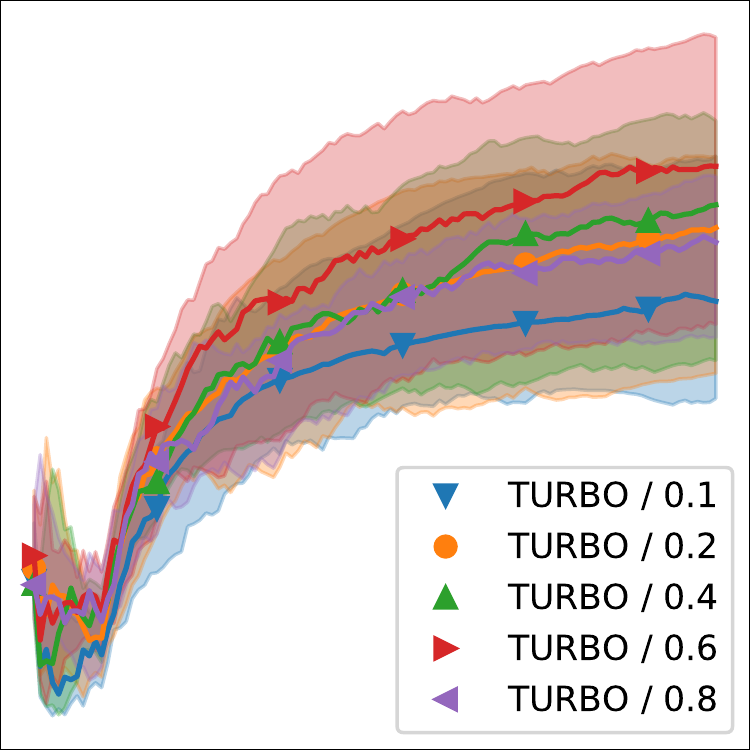};
 \end{axis}
 \end{tikzpicture}
\end{minipage}\hfill
\begin{minipage}[t]{0.33\linewidth}
 \begin{tikzpicture}
 \begin{axis}[
 x label style={at={(0.5,0.0)},anchor=center},
xmin=-5.4,xmax=113.4,
ymin=-2.5e+01,ymax=7.5e+01,
 xlabel=\# episodes,
 ylabel=Avg. observed return $\avgreward$,
 y label style={at={(0.10,0.5)},anchor=center},
 major tick length=0.1cm,
 minor tick length = 0.05cm,
 tick pos=left,
 height=\heightopenai,
 width=1.0\linewidth,
 axis on top,
 max space between ticks=20
 ]
 \addplot[] graphics[xmin=-5.4,xmax=113.4, ymin=-2.5e+01,ymax=7.5e+01,] {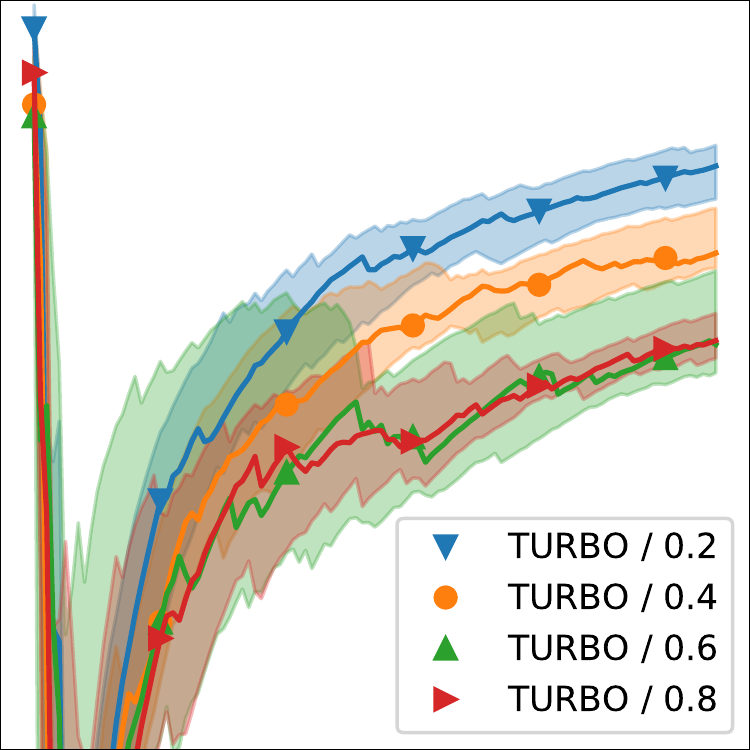};
 \end{axis}
 \end{tikzpicture}
\end{minipage}
  \caption{Results for OpenAI gym environments using different values for each method's exploration parameters. Left column: \texttt{Pendulum-v0}, center column: \texttt{CartPole-v1}, right column: \texttt{MountainCarContinuous-v0}. Top row: \gls{acr:crbo}, center row: CMA-ES \citep{Hansen2001CMAES}, bottom row: TuRBO~\citep{Eriksson2019TuRBO}.}
  \label{fig:openai_supplementary}
\end{figure}

\paragraph{Furuta Pendulum}
See \lffigref{fig:furuta_supplementary} for the results.
As expected, larger exploration, i.e., higher values for the initial standard deviation $\sigma_0$ of CMA-ES' search distribution, leads to faster convergence (right figure) at the cost of smaller average return in the beginning (left figure).
For large exploration ($\sigma_0 = 0.1$, green curve), the pendulum oftentimes became unstable and was on the brink of damaging the system.
Therefore, the full experiment in the main paper is conducted with $\sigma_0 = 0.05$, which results in better average return at the cost of slightly smaller final performance.


\begin{figure}[]
 \figurefontsize
 \centering
\begin{minipage}[t]{0.5\linewidth}
\begin{tikzpicture}
\begin{axis}[
x label style={at={(0.5,0.0)},anchor=center},
xmin=-3.2,xmax=67.2,
ymin=4.652e-01,ymax=7.436e-01,
xlabel=\# episodes,
ylabel=Avg. observed return $\avgreward$,
y label style={at={(0.05 ,0.5)},anchor=center},
major tick length=0.1cm,
minor tick length = 0.05cm,
tick pos=left,
height=0.8\linewidth,
width=1.0\linewidth,
axis on top,
max space between ticks=20
]
\addplot[] graphics[xmin=-3.2,xmax=67.2,ymin=4.652e-01,ymax=7.436e-01,,] {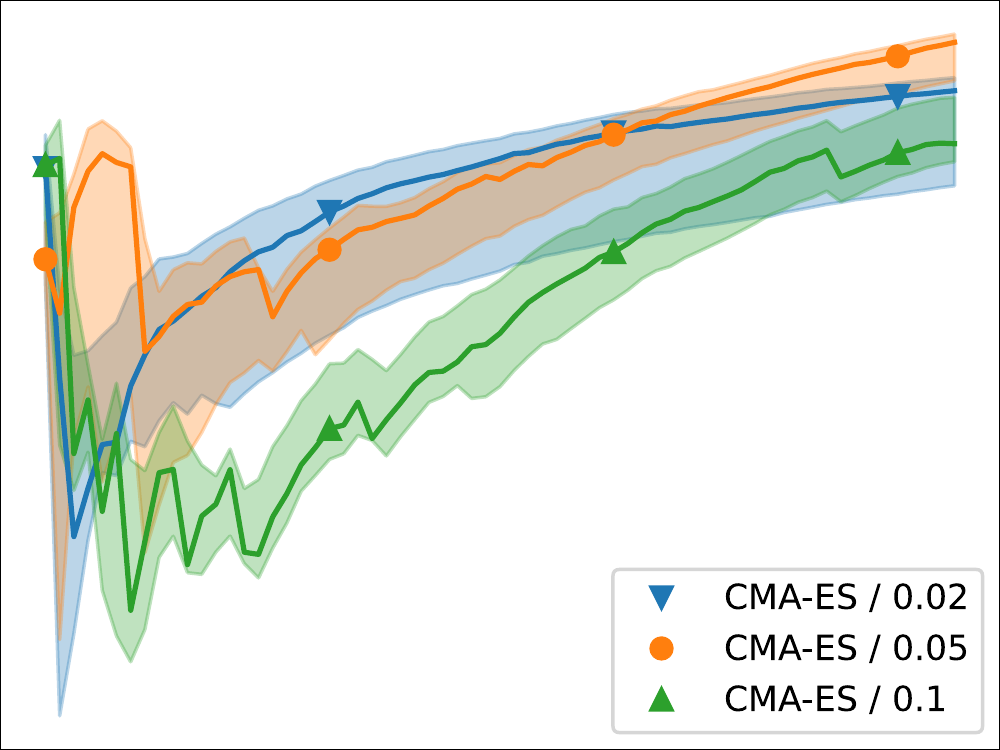};
\end{axis}
\end{tikzpicture}
\end{minipage}\hfill
\begin{minipage}[t]{0.5\linewidth}
\begin{tikzpicture}
\begin{axis}[
x label style={at={(0.5,0.0)},anchor=center},
xmin=-3.2,xmax=67.2,
ymin=6.269e-01,ymax=8.662e-01,
xlabel=\# episodes,
ylabel=Best observed return,
y label style={at={(0.05 ,0.5)},anchor=center},
major tick length=0.1cm,
minor tick length = 0.05cm,
tick pos=left,
height=0.8\linewidth,
width=1.0\linewidth,
axis on top,
max space between ticks=20
]
\addplot[] graphics[xmin=-3.2,xmax=67.2,ymin=6.269e-01,ymax=8.662e-01,] {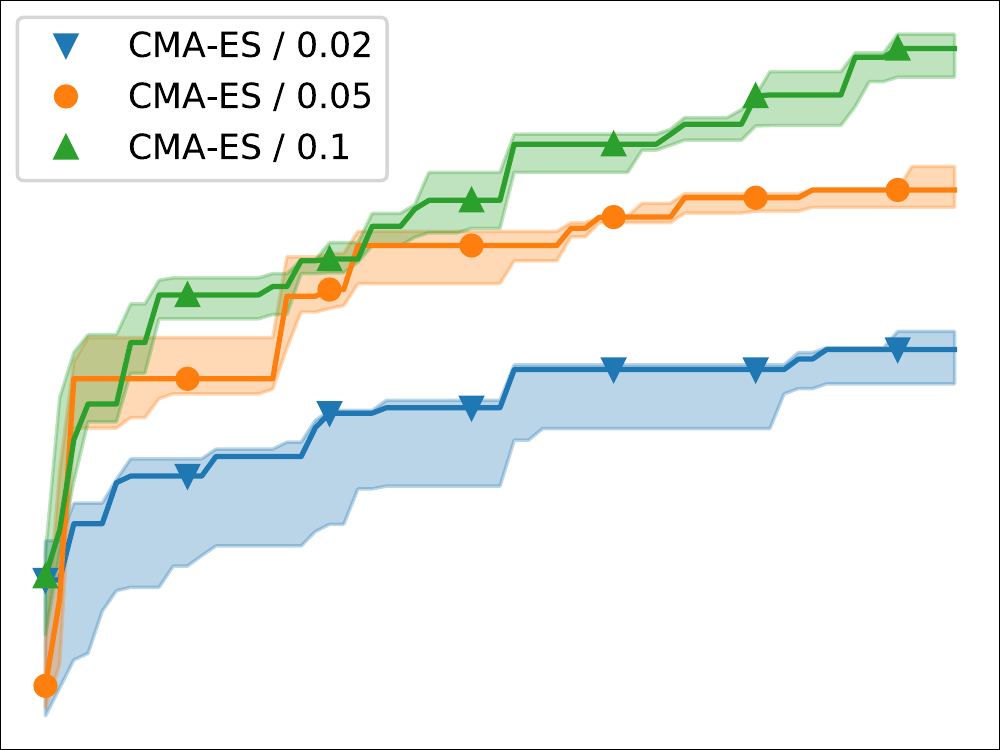};
\end{axis}
\end{tikzpicture}
\end{minipage}
\caption{Results for the experiment on the Furuta pendulum comparing different values for the initial standard deviation of CMA-ES' search distribution. Left: Average observed return across all episodes, right: best observed return during optimization.}
\label{fig:furuta_supplementary}
\end{figure}

\subsection{State Trajectories During Learning Process}
The reward function for the Furuta pendulum experiment encodes accurate tracking of the sinusoidal reference trajectory for the angle of the pendulum's horizontal arm.
In \lffigref{fig:furuta} we show that \gls{acr:crbo} outperforms CMA-ES both in terms of the average performance as well as the final performance.
In \lffigref{fig:furuta_trajectories} we visualize the resulting state trajectories for policies at different stages during the learning process for both methods.


\begin{figure}[]
 \figurefontsize
 \centering
\begin{minipage}{\linewidth}
\begin{tikzpicture}
\begin{axis}[
x label style={at={(0.5,0.0)},anchor=center},
xmin=-1.0,xmax=20.996,
ymin=-80.0,ymax=80.0,
xlabel=Time -- seconds,
ylabel=Horizontal arm angle -- degrees,
y label style={at={(0.02 ,0.5)},anchor=center},
major tick length=0.1cm,
minor tick length = 0.05cm,
tick pos=left,
height=0.4\linewidth,
width=\linewidth,
axis on top,
max space between ticks=20
]
\addplot[] graphics[xmin=-1.0,xmax=20.996,ymin=-80.0,ymax=80.0] {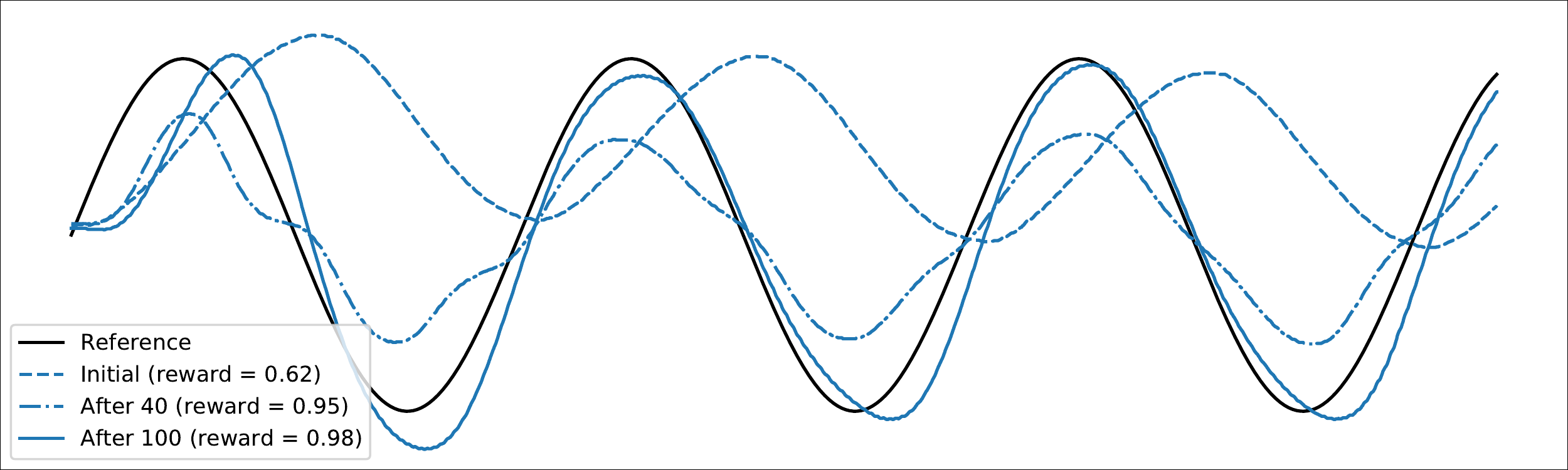};
\end{axis}
\end{tikzpicture}
\end{minipage}
\begin{minipage}{\linewidth}
\begin{tikzpicture}
\begin{axis}[
x label style={at={(0.5,0.0)},anchor=center},
xmin=-1.0,xmax=20.996,
ymin=-80.0,ymax=80.0,
xlabel=Time -- seconds,
ylabel=Horizontal arm angle -- degrees,
y label style={at={(0.02 ,0.5)},anchor=center},
major tick length=0.1cm,
minor tick length = 0.05cm,
tick pos=left,
height=0.4\linewidth,
width=\linewidth,
axis on top,
max space between ticks=20
]
\addplot[] graphics[xmin=-1.0,xmax=20.996,ymin=-80.0,ymax=80.0] {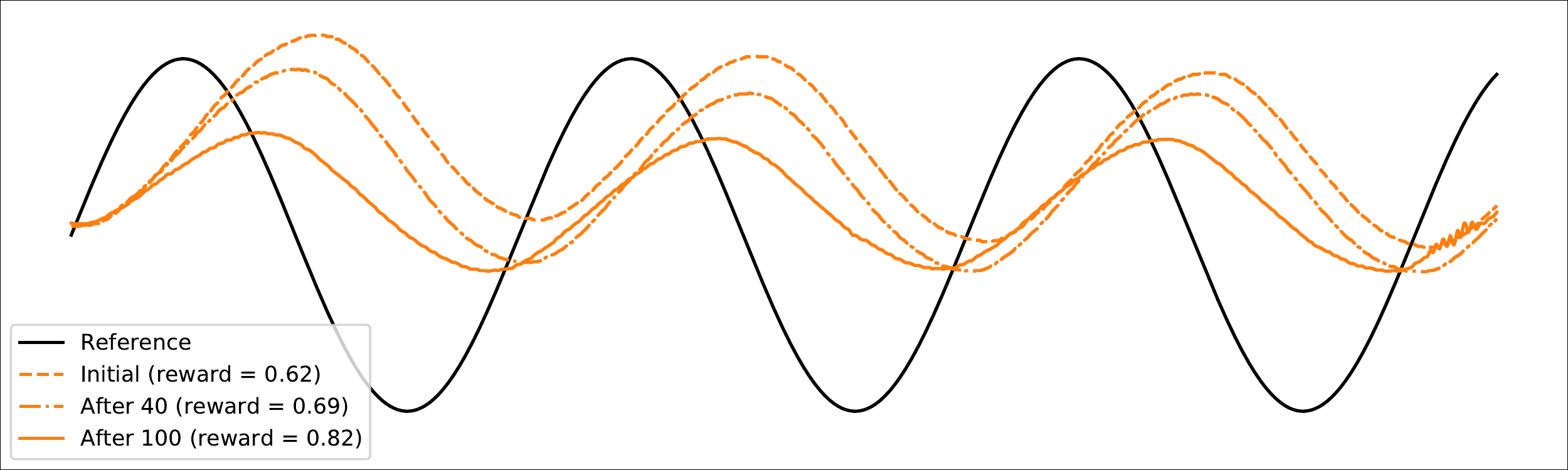};
\end{axis}
\end{tikzpicture}
\end{minipage}
\caption{State trajectories at different stages during learning progress for CRBO (top, blue) and CMA-ES (bottom, orange). The resulting trajectory for the initial policy $\pi_0$ is depicted by the dashed lines and the reference trajectory as solid black line. Trajectories after 40 and 100 iterations are shown as dashed-dotted and colored solid lines, respectively.}
\label{fig:furuta_trajectories}
\end{figure}

\section{Experimental Details}\label{app:experimental_details}

\subsection{Methods}

For each method, the policy parameters $\btheta$ are mapped from the enclosing domain $\bm{\Theta}$ (specified by lower and upper bounds, $\btheta_l, \btheta_u$) to the unit box $[0, 1]^d$.
If applicable, the exploration parameters are specified relative to the unit box.
All \gls{acr:bo}-based methods use the Mat\'ern~5/2 kernel.

\subsubsection{\Gls{acr:crbo}}

\begin{itemize}
 \item Exploration parameter: $\gamma$, defining the effective size of the confidence region, see Equation~(3) in the main paper.
 \item Acquisition function: Upper confidence bound (UCB) $\alpha(\btheta) = \mu(\btheta) + \beta \sigma(\btheta)$ with $\beta = 2.0$.
 \item Similar to Local~\gls{acr:bo}, we observed that centering the observed function values influences the exploration behavior. Therefore, we subtract the maximum of the observed function values from the dataset, i.e., $\tilde{y}_i = y_i - \max_{j=1:n} y_j$, which results in more optimistic exploration (as opposed to Local~\gls{acr:bo} where the minimum is subtracted resulting in more pessimistic exploration, \citep{Akrour2017LocalBO}).
 \item Number of samples used for Hit-and-Run sampler: $K = 1000$
 \item Number of points for initial design (see also \lfsecref{sec:initial_design}): $n_{\text{init}} = \ceil{\sqrt{\dim \btheta}}$
\end{itemize}

\subsubsection{CMA-ES \citep{Hansen2001CMAES, Hansen2019pycma}}

\begin{itemize}
 \item Exploration parameter: $\sigma_0$, initial standard deviation of the search distribution.
 \item We set the mean of the initial search distribution to the initial policy, $\mu_0 = \btheta_0$, and additionally added the initial policy as first evaluation for CMA-ES.
 \item All remaining parameters are kept to their respective default values.
\end{itemize}

\subsubsection{TuRBO \citep{Eriksson2019TuRBO}}

\begin{itemize}
 \item Exploration parameter: $\ell_0$, initial side length of the rectangular trust-region.
 \item We employ the TuRBO-1 variant, i.e., only one trust region is initialized, with batch size 1
 \item Number of points for initial design: $n_{\text{init}} = \ceil{\sqrt{\dim \btheta}}$
 \item The initial policy for all experiments was typically chosen as the center of the search domain. As TuRBO uses Latin hypercube sampling (not the Sobol sequence) for the initial design, we modified the code such that the initial policy was evaluated first.
 \item All remaining parameters are kept to their respective default values.
\end{itemize}

\subsubsection{Local BO \citep{Akrour2017LocalBO}}

\begin{itemize}
 \item Exploration parameter: $\sigma_0$, initial standard deviation of the search distribution.
 \item Number of samples per iteration: $10$ (as in the original paper)
 \item KL-constraint between updates: $1.0$ (as in the original paper)
 \item Entropy reduction constraint: $1.0$ (as in the original paper)
 \item We use the same priors for the \gls{acr:gp} hyperparameters as for \gls{acr:crbo}.
\end{itemize}

\subsection{Experiments}\label{app:experiment_details}

\subsubsection{Learning from Demonstrations: Basketball Task}
The policy vector $\btheta$ defines the DMP's forcing terms (15 each) as well as the goal attractor angle (1 each) for each degree of freedom.
Consequently, the full policy vector has $7 \times (15 + 1) = 112$ parameters.
The initial policy vector is denoted by $\btheta_0$.
\begin{itemize}
 \item Domain for forcing terms: $\bm{\Theta}_i = [\btheta_{0, i} - 200, \btheta_{0, i} + 200]$ for all $i$ corresponding to forcing terms in $\btheta$.
 \item Domain for attractor angle: $\bm{\Theta}_j = [\btheta_{0, j} - 20^\circ, \btheta_{0, j} + 20^\circ]$ for all $j$ corresponding to attractor angle terms in $\btheta$.
\end{itemize}

The exploration parameters for each method is summarized in the following table. See Figure~4 in the main paper for the corresponding results.

\begin{table}[h!]
 \centering
 \begin{tabular}{|c c||c c c |}
  \hline
  Method      & Parameter  & Small & Moderate & Large \\ \hline\hline
  CRBO (ours) & $\gamma$   & 0.4   & 0.6      & 0.9   \\ 
  CMA-ES      & $\sigma_0$ & 0.01  & 0.05     & 0.1   \\ 
  TuRBO       & $\ell_0$   & 0.05  & 0.2      & 0.8   \\ 
  Local BO    & $\sigma_0$ & 0.025 & 0.05     & 0.1   \\ \hline
  
 \end{tabular}
\end{table}

\subsubsection{Adapting Deep RL Agents to New Rewards Signals}

The policy vectors $\btheta$ corresponds to the weights of the last layer of the respective \gls{acr:nn}.
The initial policy vector is denoted by $\btheta_0$ and we chose the following domains for each of the three environments:
\begin{itemize}
 \item Pendulum-v0: $\bm{\Theta}_i = [\btheta_{0, j} - 0.5, \btheta_{0, j} + 0.5]$ for all $i$.
 \item CartPole-v1: $\bm{\Theta}_i = [\btheta_{0, j} - 10.0, \btheta_{0, j} + 10.0]$ for all $i$.
 \item MountainCarContinuous-v0: $\bm{\Theta}_i = [\btheta_{0, j} - 1.0, \btheta_{0, j} + 1.0]$ for all $i$.
\end{itemize}

The exploration parameters for the results presented in the main paper were chosen to maximize average observed reward after optimization.
In particular, the following values were used to create the results for Figure~5:

\begin{table}[h!]
 \centering
 \begin{tabular}{|c c||c c c |}
  \hline
  Method      & Parameter  & Pendulum-v0 & CartPole-v1 & MountainCar-v0 \\ \hline\hline
  CRBO (ours) & $\gamma$   & 0.3   & 0.6   & 0.5  \\ 
  CMA-ES      & $\sigma_0$ & 0.02  & 0.02  & 0.05 \\ 
  TuRBO       & $\ell_0$   & 0.1   & 0.1   & 0.2  \\ \hline
 \end{tabular}
\end{table}

\subsubsection{Sim-to-Real: Stabilizing a Furuta Pendulum }

For the hardware experiment, we use the Quanser Qube Servo~2\footnote{\url{https://www.quanser.com/products/qube-servo-2/}} and the recently presented Python interface \citep{Polzounov2019QubeInterface}.
The initial policy vector is denoted by $\btheta_0$ and we chose the following domain,  $\bm{\Theta}_i = [\btheta_{0, j} - 0.7, \btheta_{0, j} + 0.7]$ for all $i$.
The exploration parameters were set to $\gamma = 0.5$ (\gls{acr:crbo}) and $\sigma_0 = 0.05$ (CMA-ES).

\section{Practical Considerations}

For \gls{acr:bo} to work efficiently and reliably, one has to consider a few important design choices, two of which are directly affected by locally constraining the search space:
the initial design as well as the choice of priors for the \gls{acr:gp} hyperparameters.

\subsection{Initial Design}\label{sec:initial_design}

In standard \gls{acr:bo}, the first evaluations are typically chosen from a low-discrepancy sequence (e.g., Sobol or Latin hypercube sampling) to initialize the \gls{acr:gp} model with maximally diverse samples.
This is clearly not a viable solution for \gls{acr:crbo}, as these sequences only work on rectangular domains.
In contrast, our goal is to learn as much as possible about the objective function close to the initial policy $\btheta_0$.
Recall from the discussion in the main paper that the initial confidence region $\mathcal{C}_0$ is defined by a ball at $\btheta_0$ whose radius $r_0$ is known.
Thus, we sample the initial points uniformly random on the surface of this ball to get a good estimate of the gradient direction around $\btheta_0$.

\subsection{Choice of Hyperpriors for GP}

With the initial design outlined above, we expect a relatively low signal-to-noise ratio in the beginning of the optimization, as opposed to a space-filling design over the full parameter space in standard \gls{acr:bo}.
As a consequence, we reflect this in our choice of the hyperpriors for the signal and noise variances, respectively.
More specifically, we choose Gamma priors for each with the following parameters:
\begin{align*}
p_\Gamma(\btheta; \alpha, \beta) & = \frac{\beta^\alpha}{\Gamma(\alpha)} \btheta^{\alpha - 1} e^{-\beta \btheta} \\
\text{Signal variance:} & \quad  \alpha = 2.0, \beta = 0.15\\
\text{Noise variance:} & \quad  \alpha = 1.1, \beta = 0.05
\end{align*}
Similarly, we choose a Gamma prior for the kernel lengthscale with $\alpha = 3.0, \beta = 6.0$.

}
{}

\end{document}